\documentclass[]{fairmeta}
\usepackage{amsmath,amssymb}
\usepackage{afterpage}
\usepackage{booktabs}
\usepackage{colortbl}
\usepackage{makecell}
\usepackage{multirow}
\usepackage{siunitx}
\usepackage{tabularx}
\usepackage{threeparttable}

\microtypesetup{expansion=false}
\newcommand{\hounsbench}{\textsc{HounsBench}}
\newcommand{\hounsworld}{\textsc{HounsWorld}}
\newcommand{\readout}{\textsc{HounsBench-Readout}}
\newcommand{\reconstruction}{\textsc{HounsBench-Reconstruction}}
\newcommand{\simulation}{\textsc{HounsBench-Simulation}}

\newcommand{\materials}{supp/materials}
\definecolor{HBBlueDark}{HTML}{2F6690}
\definecolor{HBPurpleDark}{HTML}{9C4F96}
\definecolor{HBBlueLight}{HTML}{E8F1F8}
\definecolor{HBPurpleLight}{HTML}{F5E7F4}
\definecolor{HBYellowLight}{HTML}{FFF5DF}
\definecolor{HBGrayLight}{HTML}{F3F5F7}
\definecolor{HBSameDark}{HTML}{6D3C69}
\definecolor{HBSimDark}{HTML}{9A6200}
\newcommand{\tbh}[1]{\textcolor{white}{\bfseries #1}}
\newcommand{\samestate}{\textcolor{HBSameDark}{\ensuremath{\boldsymbol{t}}}}
\newcommand{\futurestate}{\textcolor{HBSimDark}{\ensuremath{\boldsymbol{t+\Delta t}}}}
\title{HounsWorld: A Multimodal World Model for\\
Hidden Patient-State Readout, Reconstruction, and Simulation}

\author[1,2,*]{Yunhao Bai}
\author[2,3,4,*,\ddagger]{Zhongwei Qiu}
\author[2,3,4]{Guangyu Guo}
\author[2]{Yiming Huang}
\author[2]{Tony C. W. Mok}
\author[2]{Qinji Yu}
\author[2]{Ling Zhang}
\author[1,\dagger]{Yan Wang}

\affiliation[1]{Shanghai Key Laboratory of Multidimensional Information
Processing, East China Normal University}
\affiliation[2]{DAMO Academy, Alibaba Group}
\affiliation[3]{Hupan Laboratory}
\affiliation[4]{Zhejiang University}

\contribution[*]{Equal contribution}
\contribution[\dagger]{Corresponding author}
\contribution[\ddagger]{Project leader}

\abstract{

Clinical intelligence requires estimating a patient's underlying condition from incomplete observations rather than learning isolated mappings from scans to answers. Volumetric medical images provide dense observations of anatomy, attenuation, and lesions, whereas clinical language provides sparse but complementary semantic observations. We formulate CT-centered intelligence as inference over a shared latent patient state, under which readout, reconstruction, and simulation all become state-dependent prediction problems. To operationalize this view, we introduce \textbf{HounsBench}, a computed tomography (CT) centric patient-state benchmark that unifies these three task families with patient-disjoint splits and per-family metrics, and \textbf{HounsWorld}, a 3B multimodal world model that treats volumetric scans and language as observations of the shared state through Joint Understanding-Generation Learning. A shared transformer forms an implicit patient-state estimate and supports three outputs: query-conditioned answers that read out the state, reports and captions that reconstruct it in language, and condition-specific CT volumes for low-dose denoising, virtual contrast enhancement, and anatomy-constrained text-and-mask-to-volume generation. Zero-initialized CT adapters preserve pretrained multimodal mappings, while condition-explicit Hounsfield-unit window sampling exposes clinically meaningful density observations. HounsWorld shows strong performance across all three task families while consistently improving CT understanding through clinically structured completion. Our project is available at \textcolor{blue}{https://github.com/byhwhite/HounsWorld.git}
}

\metadata[Email]{\email{yhbai@stu.ecnu.edu.cn}, \email{qiuzhongwei.qzw@alibaba-inc.com}}
\date{\today}

\begin{document}
\thispagestyle{firstheader}
\maketitle
\pagestyle{empty}

% Fill the remainder of page 1 with the introduction, then insert the overview
% float at the top of the next page.
\afterpage{%
  \begin{figure}[t!]
      \centering
      \includegraphics[width=0.95\textwidth]{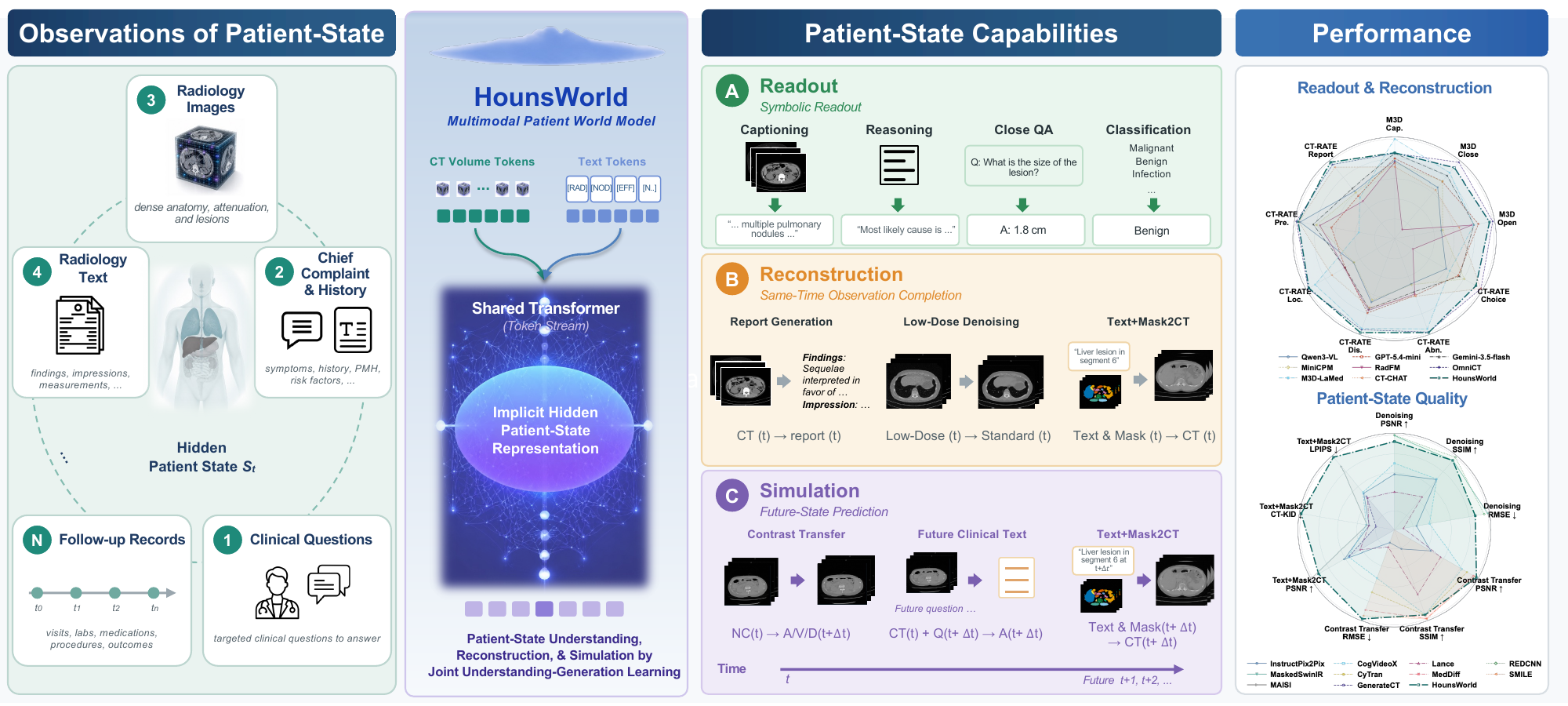}
      \caption{HounsWorld learns a shared latent representation of patient state
      from multimodal observations, under which clinical readout,
      reconstruction, and simulation can all be formulated as state-dependent
      prediction problems.}
      \label{fig:overview}
  \end{figure}%
}

\section{Introduction}

Recent progress in artificial intelligence has led to rapid adoption of learning-based systems in healthcare. However, medical intelligence remains difficult to scale because the domain is both task-diverse and data-heterogeneous: real-world clinical workflows involve many scenarios, and relevant patient information is distributed across images, text, structured variables, and annotations. This complexity makes unified modeling challenging and has contributed to the dominance of narrow task-specific solutions.

Most existing medical AI systems still address this challenge through task-specific formulations. In current practice, report generation, VQA, classification, denoising, phase translation, and text- or mask-guided synthesis are usually developed as separate problems, often with different datasets, objectives, and model interfaces. Even recent multimodal and volumetric vision--language models largely inherit this organization \cite{li2023llavamed,moor2023medflamingo,chen2024huatuogptvision,lin2025healthgpt,wuchaoyi2025radfm,hamamci2024ctrate,bai2024m3d,shi2024med2e3,gai2025threedrad,lin2026omnict}: understanding is supervised mainly through language, while image reconstruction or synthesis is treated as an auxiliary or independent generation task. As a result, supervision remains fragmented, and models are not explicitly encouraged to build a shared patient-specific representation that transfers across tasks and conditions. Reconstructive objectives can provide denser constraints and more transferable features \cite{he2022mae,li2023mage,yang2023diffusionrep,cheng2023prior,tu2026mdae}, but in medical multimodal systems they are usually optimized as separate synthesis objectives rather than as complementary observations of the same patient.

World models offer an appealing alternative because they learn latent states that explain multiple observations and support prediction under varying conditions \cite{ha2018world,hafner2025dreamer,bruce2024genie,hu2023gaia}. This perspective has been successful in sequential decision-making and generative modeling, where a shared internal state enables prediction, simulation, and control across diverse settings. The same abstraction is especially natural in medicine, because clinical intelligence is fundamentally a latent-state inference problem. A patient is not directly observed; instead, clinicians reason from partial evidence about anatomy, tissue density, and enhancement behavior. CT, language, masks, and acquisition conditions should therefore be viewed not as unrelated modalities, but as complementary observations of the same hidden patient state.

Motivated by this view, as shown in Fig.~\ref{fig:overview}, we propose \textbf{HounsWorld}, a CT-centered multimodal world model designed to reason over and generate 3D CT volumes. HounsWorld uses a shared causal transformer to integrate volumetric CT, language, anatomy masks, and condition descriptors into a unified patient-state workspace. From this shared representation, through joint understanding-generation learning, the model supports both understanding and generation: it can answer clinical questions, generate reports or captions, and complete condition-specific CT volumes for LDCT denoising, virtual contrast enhancement, and text-and-mask-conditioned generation. To make this feasible within pretrained multimodal interfaces, we further introduce clinically structured patient-state completion, condition-explicit CT observations, pseudo-frame construction for volumetric encoding, and branch-specific residual adapters for CT adaptation.

To systematically study this formulation, we construct \textbf{HounsBench}, a CT-centered benchmark that regards various tasks as a single state-dependent prediction problem. HounsBench organizes tasks into three unified families, namely symbolic and open-ended readout, same-state reconstruction, and condition-shifted simulation, providing patient-disjoint train and test splits together with per-family protocols and metrics. On HounsBench, we report unified and task-wise performance, showing that a single HounsWorld remains competitive across all three families.Also, the frozen HounsWorld reward model improves an independent Qwen3-VL-4B on open-ended QA despite closed-form-only rewards, indicating that the learned patient-state structure transfers beyond its own architecture.
Our contributions are threefold:
\begin{itemize}
\item We formulate CT-centered clinical intelligence as multimodal world modeling, casting various CT-centered tasks as a single state-dependent prediction problem, and introduce \textbf{HounsWorld}, a 3B multimodal world model that unifies readout, reconstruction, and simulation within pretrained multimodal interfaces through Joint Understanding-Generation Learning.
\item We construct \textbf{HounsBench}, a CT-centered patient-state benchmark that establishes three unified evaluation families, with patient-disjoint train and test splits and per-family protocols and metrics.
\item On HounsBench, task-wise and partial-observation analyses show that HounsWorld is competitive in readout and effective across reconstruction and simulation. Moreover, HounsWorld can also serve as a reward model for training external clinical reasoning models.
\end{itemize}

\begin{figure*}[t]
\centering
    \includegraphics[width=\textwidth]{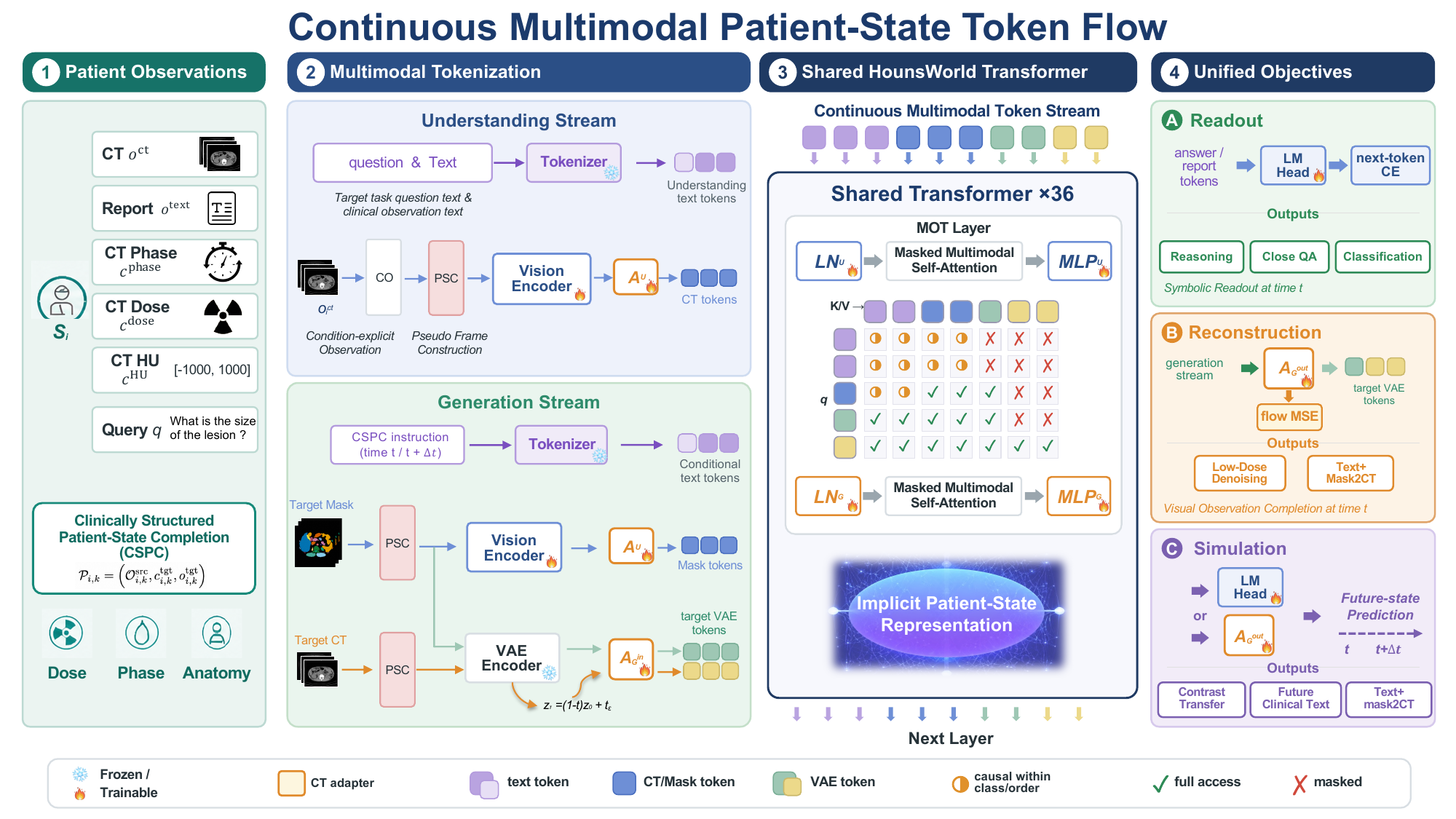}
    \caption{
    HounsWorld learns from complementary observations of a latent patient state. Clinically Structured Patient-State Completion preserves patient correspondence while changing the current or requested subsequent observation condition, jointly constraining clinical readout, text completion, and CT completion.
    }
\label{fig:framework}
\end{figure*}

\section{Related Work}
\noindent
\textbf{CT and Medical Multimodal Understanding.}
Medical VLMs adapt general multimodal models to biomedical instruction following, few-shot reasoning, and joint comprehension--generation \cite{li2023llavamed,moor2023medflamingo,chen2024huatuogptvision,lin2025healthgpt}. Radiology models extend this paradigm across 2D and 3D scans \cite{wuchaoyi2025radfm}, while CT-RATE, M3D, RadGenome, and 3D-RAD provide volumetric image--text data for report generation, VQA, grounding, and diagnostic reasoning \cite{hamamci2024ctrate,bai2024m3d,zhang2024radgenome,gai2025threedrad}. CT-specific architectures further reuse 2D backbones or compose adjacent slices to reduce volumetric cost \cite{chen2024med3dinsight,shi2024med2e3,lin2026omnict}. This line establishes that one CT can support reports, captions, localized descriptions, and explicit answers. Nevertheless, these outputs are normally organized as task formats rather than complementary observations of an underlying patient state. %HounsWorld asks whether their shared representation becomes more complete when language readout is coupled with dense CT reconstruction.

\noindent
\textbf{Medical Reconstruction and Representation Learning.}
Reconstruction objectives learn visual representations by retaining information that discriminative labels omit. Masked autoencoding, generative encoding, and diffusion features transfer effectively to recognition \cite{he2022mae,li2023mage,yang2023diffusionrep}. Medical approaches extend this idea through image--report prototype learning, representation recovery, and masked diffusion reconstruction \cite{cheng2023prior,yan2023reprec,tu2026mdae}. CT transformations provide particularly structured supervision: low-dose denoising separates acquisition noise from anatomy \cite{chen2017redcnn}, while contrast transfer predicts anatomy-dependent vascular and parenchymal appearance \cite{liu2022virtualcontrast}. These transformations preserve patient correspondence while changing how the anatomy is observed, making them natural constraints on a latent state. Prior work usually evaluates them as reconstruction systems or standalone pretraining.%; we incorporate them into a CT language model and measure transfer with matched understanding variants.

\noindent
\textbf{Multimodal Medical World Models.}
World models encode latent states from which condition-dependent observations can be predicted \cite{ha2018world,hafner2025dreamer,bruce2024genie,hu2023gaia}. Unified multimodal models bring related ideas to shared understanding and generation through common sequence backbones or aligned visual representations \cite{kondratyuk2024videopoet,wu2025janus,xie2025showo,wu2025harmon,deng2025bagel,yan2026autoencoders,fu2026lance}. Medical world modeling has used radiograph structure, tumor evolution, or radiologist gaze trajectories as the modeled state or process \cite{yue2025chexworld,yang2025mewm,li2026gazeworld}. CheXWorld is the closest conceptual precedent because it treats local anatomy, global layout, and acquisition variation as image-world structure. HounsWorld adds volumetric CT and language as complementary observation channels. Reports reconstruct the state at a semantic resolution, questions request selected facts, and clinically conditioned generation reconstructs the current scan or predicts a condition-specified subsequent observation. We study this bounded form of future-observation prediction rather than unrestricted longitudinal progression.

\section{Method}
% We study \emph{patient-state} readout, reconstruction, and simulation through Joint Understanding-Generation Learning. Volumetric CT, language, anatomy masks, and imaging conditions are treated as complementary observations of a shared latent patient state. A transformer integrates the available observations into an implicit workspace, and HounsWorld reads out queried properties, reconstructs missing observations, or predicts a requested subsequent observation.

\subsection{Patient State and Conditional Observations}

Let $s_{i,t}$ denote the latent state of patient $i$ at clinical time $t$, including volumetric anatomy, tissue attenuation, lesion manifestation, and enhancement behavior. This state is not directly supervised. Instead, the patient-aligned observations and descriptors in Fig.~\ref{fig:framework} provide partial views:
\begin{equation}
\mathcal{O}_{i,t}=
\big\{\,
\underbrace{o_{i,t}^{\mathrm{ct}},\,o_{i,t}^{\mathrm{text}}}_{\text{observations}},\;
\underbrace{c_{i,t}^{\mathrm{phase}},\,c_{i,t}^{\mathrm{dose}},\,c_{i,t}^{\mathrm{HU}}}_{\text{conditions}},\;
\underbrace{q_{i,t}}_{\text{query}}
\,\big\}.
\label{eq:observation}
\end{equation}
Here, $o^{\mathrm{ct}}$ is a CT volume and $o^{\mathrm{text}}$ is a report; $c^{\mathrm{phase}}$, $c^{\mathrm{dose}}$, and $c^{\mathrm{HU}}$ specify the contrast phase, acquisition dose, and HU window under which the CT is observed; and $q$ is a query. CT provides dense spatial and attenuation evidence, language selects clinically salient properties, and structured descriptors identify how an observation was acquired or presented.

Given source observations $\mathcal{O}_{i,\leq t}^{\mathrm{src}}$ and a requested target condition $c_{i,t+\Delta t}^{\mathrm{tgt}}$, the shared transformer forms
\begin{equation}
h_{i,t}=F_{\theta}
\left(
\mathcal{O}_{i,\leq t}^{\mathrm{src}},
c_{i,t+\Delta t}^{\mathrm{tgt}}
\right),
\end{equation}
and either completes a target observation or answers a query:
\begin{equation}
\begin{aligned}
\hat{o}_{i,t+\Delta t}^{\mathrm{tgt}}
&\sim p_{\theta}\!\left(
o_{i,t+\Delta t}^{\mathrm{tgt}}
\mid h_{i,t},c_{i,t+\Delta t}^{\mathrm{tgt}}
\right),\\
a_{i,t+\Delta t}
&\sim p_{\theta}(a\mid h_{i,t},q_{i,t}).
\end{aligned}
\label{eq:patient_world}
\end{equation}
For $\Delta t=0$, this represents same-state completion, such as report reconstruction or denoising. For a specified subsequent condition, it represents bounded future-observation prediction: arterial/venous CT follows a non-contrast observation, and future-oriented VQA requests a prospective clinical readout. This operational definition does not assume unrestricted longitudinal disease forecasting. Closed-form answers are symbolic readouts, whereas open answers, reports, and captions are language observations generated from the same workspace.

\subsection{Clinically Structured Patient-State Completion}
HounsWorld is trained through \emph{Joint Understanding-Generation Learning}, as illustrated in Fig.~\ref{fig:framework}, a single objective that couples language readout with conditional CT generation over shared patient-state observations. Rather than training separate understanding and generation heads. We instantiate this objective as \emph{Clinically Structured Patient-State Completion} (CSPC), each patient-aligned completion example is:
\begin{equation}
\mathcal{P}_{i,k}=
\left(
\mathcal{O}_{i,k}^{\mathrm{src}},
c_{i,k}^{\mathrm{tgt}},
o_{i,k}^{\mathrm{tgt}}
\right),
\label{eq:cspc_tuple}
\end{equation}
where $k$ indexes a completion objective, $\mathcal{O}_{i,k}^{\mathrm{src}}$ is the available evidence, $c_{i,k}^{\mathrm{tgt}}$ specifies the requested observation, and $o_{i,k}^{\mathrm{tgt}}$ is its patient-matched target. The three CT objectives are LDCT denoising (dose-conditioned reconstruction), virtual contrast enhancement (phase-conditioned short-horizon prediction), and text/mask-conditioned CT generation (structured conditional simulation). CT-to-report and CT-to-caption examples reconstruct sparse language observations, while VQA uses the same representation for query-conditioned readout.

\noindent\textbf{Condition-explicit observations.}
CT appearance depends jointly on patient content and the acquisition or display condition. We represent a CT condition by
\begin{equation}
c_i^{\mathrm{ct}}=
\left(
c_i^{\mathrm{dose}},
c_i^{\mathrm{phase}},
c_i^{\mathrm{HU}}
\right),
\qquad
\bar{o}_i^{\mathrm{ct}}=
\left(
\mathcal{T}_{c_i^{\mathrm{ct}}}(o_i^{\mathrm{ct}}),
\tau(c_i^{\mathrm{ct}})
\right),
\label{eq:condition_observation}
\end{equation}
where $\mathcal{T}_{c}$ produces the visual observation and $\tau(c)$ serializes the same condition in language. Intensity transformations are therefore presented together with the condition that produced them, reducing ambiguity between a change in patient content and a change in acquisition or display.

\noindent\textbf{HU-aware observations.}
For HU-window bounds $(l_w,u_w)$, the observation operator is
\begin{equation}
\mathcal{T}_{w}(V)=
\operatorname{clip}
\left(
\frac{V-l_w}{u_w-l_w},0,1
\right).
\label{eq:window}
\end{equation}
Sampling such condition-explicit windows requires the representation to associate density-selective appearance with its clinical context. During training, with probability \(\lambda_{\mathrm{CO}}\), we modify \(c^{\mathrm{HU}}\) using multiple CT observation HU values.

\noindent\textbf{Pseudo-frame construction.}
Pseudo-Frame Construction (PSC) exposes volumetric CT to pretrained RGB interfaces by packing adjacent axial slices into color channels. For a depth-$D$ volume $V_i\in\mathbb{R}^{D\times H\times W}$, let $K=\lceil D/3\rceil$ and let $\widetilde V_i$ denote its depth-boundary-padded form with $3K$ slices. PSC constructs
\begin{equation}
F_{i,k}(x,y,r)=
\widetilde V_i(3k+r,x,y),
\quad
\substack{k=0,\ldots,K-1,\\r\in\{0,1,2\}}.
\label{eq:pseudo_frame}
\end{equation}
Thus, each $F_{i,k}\in\mathbb{R}^{H\times W\times3}$ contains three adjacent slices; no valid CT slice is reused, and padding is excluded after reconstruction. PSC preserves local through-plane context while reducing sequence length by approximately threefold and provides a common interface for the vision and VAE encoders.

\noindent\textbf{CT residual adaptation.}
The semantic vision encoder and VAE pathway produce continuous tokens with different pretrained distributions. HounsWorld inserts branch-specific residual adapters after the understanding encoder, after the VAE-to-transformer input projection, and before the transformer-to-VAE output projection. For token $u\in\mathbb{R}^{d}$ in branch $b\in\{U,G_{\mathrm{in}},G_{\mathrm{out}}\}$,
\begin{equation}
\mathcal{A}_{b}(u)=
u+
W_{b,\mathrm{up}}\,
\phi
\left(
W_{b,\mathrm{down}}\,
\mathrm{RMSNorm}(u)
\right),
\label{eq:adapter}
\end{equation}
where $\phi$ is GELU and the two projections form a bottleneck. Zero-initializing $W_{b,\mathrm{up}}$ makes each adapter an identity mapping at initialization, preserving the pretrained semantic and generative interfaces while allowing branch-specific CT density and spatial corrections.

\subsection{Readout, Reconstruction, and Simulation}

HounsWorld combines language readout, text reconstruction, and clinically structured CT completion in one objective.

\noindent\textbf{Language readout and reconstruction.}
Let $\mathcal{I}_{\mathrm{read}}$ index answer tokens and $\mathcal{I}_{\mathrm{text}}$ index report/caption tokens, with $\mathcal{I}_{\mathrm{lang}}=\mathcal{I}_{\mathrm{read}}\cup\mathcal{I}_{\mathrm{text}}$. Their shared autoregressive loss is
\begin{equation}
\mathcal{L}_{\mathrm{lang}}=-
\sum_{j\in\mathcal{I}_{\mathrm{lang}}}
\log p_{\theta}(w_j\mid\mathcal{O}_i,w_{<j}).
\label{eq:language}
\end{equation}

\noindent\textbf{CT reconstruction and conditional simulation.}
CT completion is performed in VAE latent space with flow matching \cite{lipman2023flow}. Given target latent $z$, Gaussian noise $\epsilon$, and flow time $\gamma$, we construct $z_{\gamma}=(1-\gamma)z+\gamma\epsilon$ with velocity target $u^{\star}=\epsilon-z$. For completion type $k$,
\begin{equation}
\mathcal{L}_{\mathrm{ct}}^{k}=
\mathbb{E}
\left[
\left\|
u_{\theta}
\left(
z_{\gamma},\gamma,
\mathcal{O}_{i,k}^{\mathrm{src}},
c_{i,k}^{\mathrm{tgt}}
\right)
-u^{\star}
\right\|_2^2
\right].
\label{eq:flow}
\end{equation}
The complete objective is
\begin{equation}
\mathcal{L}_{\mathrm{HW}}=
\mathcal{L}_{\mathrm{lang}}
+\lambda_{\mathrm{ct}}
\sum_{k\in\mathcal{K}}
\pi_k\mathcal{L}_{\mathrm{ct}}^{k},
\label{eq:total}
\end{equation}
where $\pi_k$ is the sampling frequency of CT-completion type $k$ and $\lambda_{\mathrm{ct}}$ balances dense latent prediction against language supervision.

\noindent\textbf{Two-stage optimization.}
Let $\Theta_{\mathcal{A}}$ collect the three CT adapters and let $\Theta_{\mathrm{fixed}}$ contain the Wan VAE and pretrained latent connectors. The trainable parameter sets are
\begin{equation}
\Theta_{\mathrm{train}}^{(1)}=
\Theta_{\mathcal{A}},
\qquad
\Theta_{\mathrm{train}}^{(2)}=
\Theta\setminus\Theta_{\mathrm{fixed}}.
\label{eq:two_stage}
\end{equation}
Stage~1 adapts only the CT interfaces. Stage~2 starts from this initialization and jointly tunes the semantic vision encoder/merger, token embeddings, shared transformer, language head, and CT adapters, while the VAE and latent connectors remain frozen. The schedule first establishes a stable CT-compatible interface and then co-adapts clinical readout and completion under the unified objective.

\section{Experiments}

\begin{table*}[t!]
\centering
\footnotesize
\setlength{\tabcolsep}{8pt}
\resizebox{\textwidth}{!}{%
\begin{tabular}{@{}lccccccccc@{}}
\toprule
\multirow{2}{*}{Method}
& \multicolumn{3}{c}{\shortstack{HounsBench-Reconstruction\\Denoising ($t$)}}
& \multicolumn{3}{c}{\shortstack{HounsBench-Simulation\\Contrast Transfer ($t+\Delta t$)}}
& \multicolumn{3}{c}{\shortstack{HounsBench-Reconstruction/Simulation\\Text+Mask-to-CT ($t$ / $t+\Delta t$)}} \\
\cmidrule(lr){2-4}\cmidrule(lr){5-7}\cmidrule(lr){8-10}
& PSNR$\uparrow$ & SSIM$\uparrow$ & RMSE$\downarrow$
& PSNR$\uparrow$ & SSIM$\uparrow$ & RMSE$\downarrow$
& PSNR$\uparrow$ & CT-KID$\downarrow$ & LPIPS$\downarrow$ \\
\midrule
InstructPix2Pix & 18.961 & 0.5518 & 0.1174 & 13.979 & 0.2030 & 0.2057 & \underline{11.547} & 1.077 & 0.6024 \\
CogVideoX       & \underline{22.731} & \underline{0.5577} & \underline{0.0739}
& \underline{25.770} & \underline{0.8963} & \underline{0.0546}
& 10.574 & \underline{0.746} & \underline{0.5878} \\
Lance           & 13.057 & 0.3649 & 0.2856 & 17.772 & 0.6824 & 0.1524 & 7.905 & 1.042 & 0.6895 \\
\midrule
\textbf{HounsWorld (Ours)} & \textbf{30.196} & \textbf{0.7631} & \textbf{0.0322}
& \textbf{31.343} & \textbf{0.8997} & \textbf{0.0295}
& \textbf{17.354} & \textbf{0.075} & \textbf{0.3056} \\
\bottomrule
\end{tabular}%
}
\vspace{-0.2cm}
% \caption{Comparison with general-purpose video-editing models across three CT-completion tasks. HounsWorld ranks first on every reported metric. Bold and underlined values denote the best and second-best results within this comparison.}
\caption{Comparison with generation models across the HounsBench-Reconstruction and HounsBench-Simulation subtasks. Bold/underline: best/second-best.}
\label{tab:generation_general}
\end{table*}

\begin{table}[t!]
\centering
\footnotesize
\setlength{\tabcolsep}{13pt}
\begin{tabular}{@{}lccc@{}}
\toprule
\textbf{Denoising} & PSNR$\uparrow$ & SSIM$\uparrow$ & RMSE$\downarrow$ \\
\midrule
RED-CNN       & \textbf{31.382} & \textbf{0.7788} & \textbf{0.0310} \\
MaskDenoising & \underline{31.198} & \underline{0.7745} & 0.0329 \\
% \midrule
\textbf{HounsWorld} & 30.196 & 0.7631 & \underline{0.0322} \\
\toprule
% \midrule
\textbf{Contrast Transfer} & PSNR$\uparrow$ & SSIM$\uparrow$ & RMSE$\downarrow$ \\
\midrule
CyTran  & 27.713 & 0.9027 & 0.0422 \\
MedDiff & 30.186 & \underline{0.9386} & 0.0328 \\
SMILE   & \underline{30.736} & \textbf{0.9429} & \underline{0.0306} \\
% \midrule
\textbf{HounsWorld} & \textbf{31.343} & 0.8997 & \textbf{0.0295} \\
\toprule
% \midrule
\textbf{Text+mask-to-CT} & PSNR$\uparrow$ & CT-KID$\downarrow$ & LPIPS$\downarrow$ \\
\midrule
MAISI      & \textbf{18.780} & \underline{0.826} & \underline{0.3506} \\
GenerateCT & 10.391 & 4.255 & 0.6613 \\
% \midrule
\textbf{HounsWorld} & \underline{17.354} & \textbf{0.075} & \textbf{0.3056} \\
\bottomrule
\end{tabular}
\caption{Compared with task-specific models on the HounsBench-Reconstruction and Simulation subtasks.}

\label{tab:gen_specific}
\end{table}

% \subsection{HounsBench Construction}
% HounsBench is divided into three task families, readout, reconstruction, and simulation, each built on CT-RATE \cite{hamamci2024ctrate} and M3D \cite{bai2024m3d} with anatomy masks from TotalSegmentator \cite{wasserthal2023totalsegmentator}. Training samples across language, denoising, enhancement, and text-and-mask are drawn with weights 60/10/10/20\%. Full metric definitions, dataset construction, and baselines are provided in the \textbf{supplementary material}.
% \begin{itemize}
%     \item \textbf{HounsBench-Readout} contains about 1.43M training samples, collected from CT-RATE and M3D, including 1.22M VQA pairs and 34K held-out samples, covering closed-form VQA, open-ended VQA, multiple-choice, caption, and reasoning readout.
%     \item \textbf{HounsBench-Reconstruction} contains 34K reconstructions, 6.9K LDCT-denoising and 188K current-state text-and-mask-to-CT training pairs (0.1K and 0.2K held-out), covering the report reconstruction, caption reconstruction, LDCT-denoising, and text-and-mask-to-CT for the same-time observation completion.
%     \item \textbf{HounsBench-Simulation} contains 7.1K contrast-transfer pairs, covering the contrast transfer for future time, text task for future clinical state prediction.
% \end{itemize}

\subsection{HounsBench Construction}
HounsBench comprises three task families: readout, reconstruction, and simulation. It is constructed from CT-RATE \cite{hamamci2024ctrate}, M3D \cite{bai2024m3d}, and private collected dataset with anatomical masks generated by TotalSegmentator \cite{wasserthal2023totalsegmentator}. During training, language, denoising, enhancement, and text-and-mask samples are drawn at a ratio of 60/10/10/20\%. Detailed dataset construction, evaluation metrics, and baseline configurations are provided in the \textbf{supplementary material}.

\begin{itemize}
    \item \textbf{HounsBench-Readout} contains approximately 1.43M training samples from CT-RATE and M3D, including 1.22M VQA pairs, with 34K samples held out for evaluation. It covers closed-form, open-ended, and multiple-choice VQA, as well as captioning and reasoning tasks.
    
    \item \textbf{HounsBench-\allowbreak Reconstruction} contains 34K language-reconstruction samples, 6.9K LDCT-denoising pairs, and 188K current-state text-and-mask-to-CT pairs. This family evaluates report and caption reconstruction, LDCT denoising, and same-time CT observation completion.
    
    \item \textbf{HounsBench-Simulation} contains 7.1K contrast-transfer pairs and future-state text-and-mask-to-CT pairs, and future-state clinical text tasks, evaluating condition-shifted CT prediction and future clinical-state prediction.
\end{itemize}

\subsection{Compared Baselines}
Understanding baselines are Qwen3-VL-4B \cite{bai2025qwen3vl}, GPT-5.4-mini \cite{openai2026gpt54mini}, Gemini-3.5-flash \cite{kavukcuoglu2026gemini35}, MiniCPM-V~4.5 \cite{yu2025minicpmv45}, RadFM \cite{wuchaoyi2025radfm}, M3D-LaMed-4B \cite{bai2024m3d}, CT-CHAT-7B \cite{hamamci2024ctrate}, OmniCT-3B \cite{lin2026omnict}, and Lance \cite{fu2026lance}. Qwen3-VL and MiniCPM use 2D slice contact sheets; medical volume models use native 3D input. CT-completion baselines include InstructPix2Pix \cite{brooks2023instructpix2pix}, CogVideoX \cite{yang2025cogvideox}, Lance \cite{fu2026lance}, RED-CNN \cite{chen2017redcnn}, MaskDenoising \cite{chen2023masked}, CyTran \cite{ristea2023cytran}, MedDiff \cite{rombach2022ldm}, SMILE \cite{liu2025smile}, MAISI \cite{guo2025maisi}, and GenerateCT \cite{hamamci2024generatect}.

\begin{table*}[t!]
\centering
\footnotesize
\setlength{\tabcolsep}{2.5pt}
\resizebox{\textwidth}{!}{%
\begin{tabular}{@{}lcccccccccccc@{}}
\toprule
\multirow{2}{*}{Model}
& \multirow{1}{*}{\shortstack{HounsBench (HB.)}}
& \multicolumn{3}{c}{HB.-Readout (M3D)}
& \multicolumn{6}{c}{HB.-Readout (CT-RATE)}
& \multirow{1}{*}{HB.-Simulation}
& \multirow{2}{*}{Avg.} \\
\cmidrule(lr){2-2}\cmidrule(lr){3-5}\cmidrule(lr){6-11}\cmidrule(lr){12-12}
& \#Params 
& Cap. & Close & Open
& Choice & Abn. & Dis. & Loc. & Pre. & Report
& Future state ($t + \Delta t$)  & \\
\midrule

\multicolumn{13}{@{}l}{%
  \textbf{General LVLMs, only understanding}} \\
\midrule
Qwen3-VL
& 4B
& 32.20 & 55.28 & 23.02
& 52.51 & 22.38 & 25.72 & 24.08 & 93.19 & 37.34
& 5.02 & 37.07 \\

GPT-5.4-mini
& --
& 33.20 & 61.32 & 41.15
& 64.44 & 28.05 & 30.15 & 29.13 & 62.54 & 33.72
& 15.66 & 39.94 \\

Gemini-3.5-flash
& --
& 35.31 & 67.50 & 38.92
& 65.82 & 26.36 & 28.45 & 29.13 & 80.16 & 39.50
& 12.04 & 42.32 \\

MiniCPM
& 9B
& 31.28 & 61.50 & 23.68
& 70.61 & 22.79 & 24.85 & 23.89 & 77.02 & 36.28
& 3.51 & 37.54 \\

\midrule
\multicolumn{13}{@{}l}{%
  \textbf{Medical LVLMs, only understanding}} \\
\midrule
RadFM
& 14B
& 31.58 & 9.62 & 37.91
& 18.53 & 27.42 & 29.10 & 29.55 & 73.50 & 29.45
& 18.14 & 30.48 \\

M3D-LaMed
& 4B
& \textbf{41.27} & 70.66 & 38.87
& 56.20 & 44.59 & 36.29 & 46.67 & 35.25 & 32.31
& 9.86 & 41.20 \\

CT-CHAT
& 8B
& 29.68 & 52.40 & 31.57
& \underline{87.53} & 28.55 & 27.89 & 36.95 & 94.37 & 55.89
& 26.34 & 47.12 \\

OmniCT
& 3B
& 35.35 & \textbf{81.90} & \textbf{48.23}
& \textbf{87.61} & \underline{46.17} & \underline{36.78}
& \textbf{50.74} & \textbf{96.94} & \underline{58.01}
& \underline{57.23} & \textbf{59.90} \\

\midrule
\multicolumn{13}{@{}l}{%
  \textbf{Multimodal CT world model, unified understanding and generation}} \\
\midrule
\textbf{HounsWorld (Ours)}
& 3B
& \underline{35.55} & \underline{76.16} & \underline{46.65}
& 82.52 & \textbf{46.45} & \textbf{37.97}
& \underline{50.27} & \underline{95.13} & \textbf{59.78}
& \textbf{59.18} & \underline{58.96} \\
\bottomrule
\end{tabular}%
}

\caption{Comparison on the HounsBench-Readout subsets (M3D and
CT-RATE) and HounsBench-Simulation. Avg. is the mean of the ten task scores. All metrics are percentages, and higher
is better. Bold and underlined values denote the best and second-best
results, respectively; ``--'' denotes an unavailable result.}
\label{tab:main_results}
\end{table*}

\subsection{Main Results of HounsWorld}

\noindent\textbf{Clinically Structured CT Completion.}
HounsWorld ranks first on all nine metrics against generation models (Table~\ref{tab:generation_general}), improving PSNR over the strongest comparator by 7.465, 5.573, and 5.807\,dB for reconstruction (LDCT denoising), simulation (virtual contrast enhancement), and text+mask2CT. The phase targets are acquired after the non-contrast observation, making enhancement a bounded test of condition-specified future-observation prediction, consistent with its placement in the simulation family. Against task-specific methods (Table~\ref{tab:gen_specific}), HounsWorld achieves the best simulation PSNR/RMSE and text+mask2CT CT-KID/LPIPS. It remains close on metrics led by dedicated models, within 1.186\,dB PSNR, 0.0157 SSIM, and 0.0012 RMSE for reconstruction (LDCT denoising), and within 1.426\,dB of the best text+mask2CT PSNR. These localized gaps are a reasonable trade-off for one shared 3B model that spans readout, reconstruction, and simulation without task-specific architectures.

\noindent\textbf{CT Readout.}
For closed-form tasks, we evaluate performance using normalized accuracy; for open-ended QA, we use a weighted combination of BLEU~\cite{papineni2002bleu}, ROUGE~\cite{lin2004rouge}, RadGraph~\cite{delbrouck2024radgraphxl}, and BioBERTScore~\cite{zhang2020bertscore,lee2020biobert}. Table~\ref{tab:main_results} shows that HounsWorld ranks second overall, only 0.931 points behind OmniCT \cite{lin2026omnict}, while uniquely unifying CT readout and completion in this comparison. It obtains the best scores for abnormality (46.45), disease (37.97), report generation (59.78) and simulation (59.18), exceeding the strongest readout-only result by 0.28, 1.19, 1.77 and 1.95 points. HounsWorld also ranks second on all three M3D tasks and on CT-RATE localization and prediction. Thus, the unified objective preserves competitive readout, with a modest aggregate gap to the strongest specialized readout model.

% \noindent\textbf{Clinically Structured CT Completion.}
% HounsWorld ranks first on all nine metrics against general-purpose editing models (Table~\ref{tab:generation_general}), improving PSNR over the strongest comparator by 7.465, 5.573, and 5.807\,dB for LDCT denoising, virtual contrast enhancement, and text-and-mask-conditioned CT generation. The phase targets are acquired after the non-contrast observation, making enhancement a bounded test of condition-specified future-observation prediction. Against task-specific methods (Table~\ref{tab:gen_specific}), HounsWorld achieves the best enhancement PSNR/RMSE and text-and-mask CT-KID/LPIPS. It remains close on metrics led by dedicated models, within 1.186\,dB PSNR, 0.0157 SSIM, and 0.0012 RMSE for LDCT denoising, and within 1.426\,dB of the best text-and-mask PSNR. These localized gaps are a reasonable trade-off for one shared 3B model that spans readout, reconstruction, and simulation without task-specific architectures.

\subsection{Ablation Study and Model Analysis}
\begin{table}[t]
\centering
\footnotesize
\setlength{\tabcolsep}{2.0pt}
\begin{tabular}{@{}lccccccc@{}}
\toprule
Training strategy & Choice & Abn. & Dis. & Loc. & Pre. & Rep. & Avg. \\
\midrule
UND & 79.08 & 45.57 & 33.48 & 49.11 & 93.29 & 58.62 & 59.86 \\
\cmidrule(lr){1-8}
UND+DEN & \underline{81.96} & 44.90 & 36.66 & 49.88 & 94.31 & 59.44 & 61.19 \\
\cmidrule(lr){1-8}
UND+VE & 81.31 & \underline{46.84} & 36.35 & 49.87 & \textbf{95.23} & 59.32 & 61.49 \\
\cmidrule(lr){1-8}
UND+T2C & 81.27 & 46.24 & 36.50 & 50.07 & 94.90 & 59.35 & 61.39 \\
\cmidrule(lr){1-8}
UND+CSPC & 81.93 & \textbf{47.26} & \underline{37.11} & \underline{50.17} & 95.02 & \underline{59.53} & \underline{61.84} \\
\cmidrule(lr){1-8}
UND+CSPC+CO & \textbf{82.52} & 46.45 & \textbf{37.97} & \textbf{50.27} & \underline{95.13} & \textbf{59.78} & \textbf{62.02} \\
\bottomrule
\end{tabular}

\captionof{table}{Effect of clinically structured CT completion on Hounsbench-Readout(CT-RATE). UND: understanding; DEN/VE/T2C: LDCT denoising, contrast transfer, and text+mask2CT; CSPC: all three; CO: condition-explicit observation.}
\label{tab:generation_ablation}
\end{table}

\begin{figure}[t]
\centering
    \includegraphics[width=0.68\linewidth]{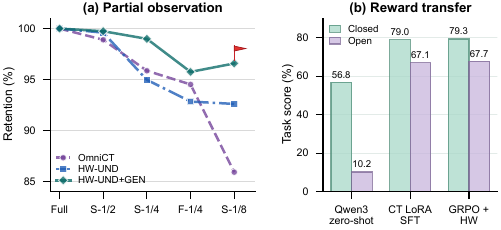}
    \caption{Complementary ablations. (a) Report-quality retention under uniformly sparse (S) or central field-of-view (F) observations; CSPC degrades most gracefully. (b) Closed- and open-task transfer to an independent Qwen3-VL policy; Reinforcement learning improves using HounsWorld (HW) as rewards over zero-shot and SFT.}
\label{fig:compact_ablations}
\end{figure}

\begin{figure*}[t!]
    \centering
    \includegraphics[width=0.99\textwidth]{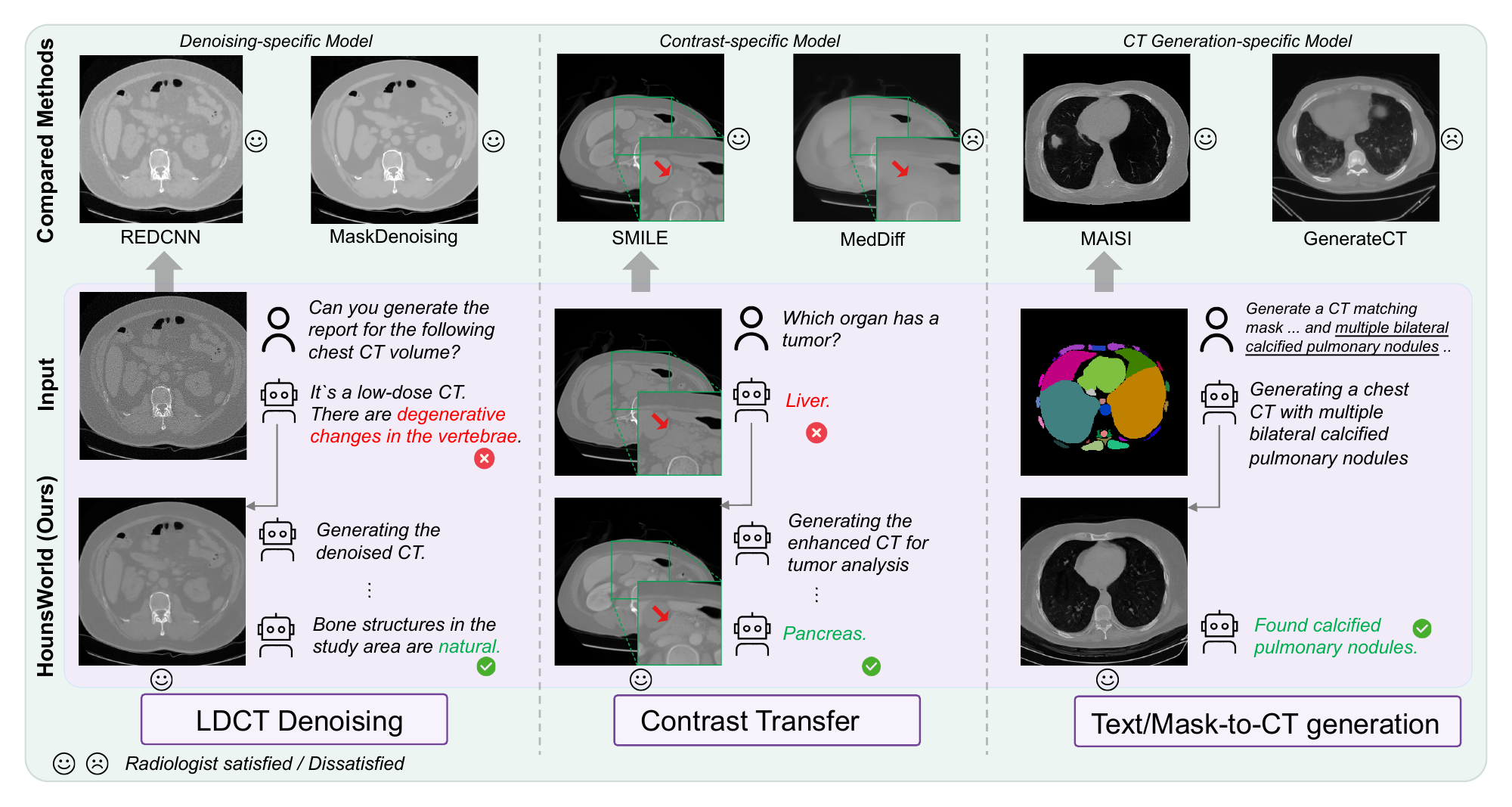}
    % \caption{Qualitative consistency across condition-specific observations. All interactions in the blue panels are HounsWorld. Left: denoising changes the inferred finding from vertebral degeneration to normal bone structure. Center: a predicted post-contrast observation separates vessels from the pancreatic lesion (red arrows) and changes localization from liver to pancreas. Right: text-and-mask conditioning instantiates multiple bilateral calcified pulmonary nodules that remain identifiable in the synthesized CT. The top row shows task-specific comparators.}
    \vspace{-0.5em}
    \caption{Qualitative consistency across condition-specific observations. The top row shows task-specific comparators. Left: denoising revises the inferred finding from vertebral degeneration to normal bone. Center: a predicted post-contrast observation separates vessels from the pancreatic lesion (red arrows) and shifts localization from liver to pancreas. Right: text-and-mask conditioning instantiates multiple bilateral calcified pulmonary nodules that remain identifiable in the synthesized CT.}
    \label{fig:generation_visualization}
\end{figure*}

% \begin{figure}[t!]
%     \centering
%     \includegraphics[width=0.90\columnwidth]{figures/transfer_reward_ablation.pdf}
%     \caption{Closed- and open-task scores of an independent Qwen3-VL-4B-Instruct policy. Both GRPO variants share data and SFT initialization; only reward shaping differs. HounsWorld shaping consistently improves both task groups.}
%     \label{fig:transfer_reward}
% \end{figure}

\noindent \textbf{Completion-to-Readout Transfer.}
Table~\ref{tab:generation_ablation} tests whether CT-completion supervision strengthens the shared representation used for understanding, rather than merely adding a generative output path. We keep the language-supervised objective and evaluation protocol fixed, initialize every variant independently from the same Lance checkpoint, and vary the added completion signal. UND is the understanding-only control; UND+DEN, UND+VE, and UND+T2C add dose completion, phase completion, or anatomy-and-language completion individually; UND+CSPC combines all three; and the final variant additionally exposes the CT acquisition condition through CO. Every model is evaluated on the same six categories, and Avg.\ is their unweighted mean.

Each individual completion objective improves the average over UND, with VE providing the strongest single-task gain (+1.63). This is consistent with phase-transition supervision coupling current anatomy to its subsequent contrast-dependent appearance. Combining the three objectives through CSPC improves every category over UND and yields a larger +1.98 gain, indicating complementary supervision across dose, phase, and anatomy/text conditions. CO adds a smaller improvement concentrated in Choice, Disease, Localization, and Report, bringing the total gain to +2.16.

% \noindent \textbf{Robustness under Partial CT Observation.}
% The full-volume ablation does not reveal whether CSPC remains useful when the observed CT is incomplete. We therefore test the hypothesis that completion supervision encourages a patient-state representation that degrades gracefully as evidence is removed. The comparison between uniform sparsification and a contiguous field-of-view crop also separates the amount of retained evidence from its anatomical coverage.

% We evaluate OmniCT-3B \cite{lin2026omnict}, HounsWorld-UND, and HounsWorld-CSPC on the same 0.5K randomly selected CT-RATE volumes. Starting from the standard full-volume input, Sparse-$r$ uniformly retains fractions $r\in\{1/2,1/4,1/8\}$ and resamples the retained slices to each model's required input length, whereas FOV-$1/4$ retains only the contiguous central quarter. Decoding and report scoring are unchanged. For each model, retention is the partial-observation report score divided by its own full-observation score, so the experiment measures relative robustness rather than absolute report quality or exact recovery of every unobserved finding.

% CSPC is most useful as uniform sampling becomes severe: at Sparse-$1/8$, it improves retention by 3.97 points over UND and 10.66 points over OmniCT-3B. Its higher retention under Sparse-$1/4$ than FOV-$1/4$ further suggests that distributed anatomical coverage is more informative than a contiguous central field of view.

\noindent \textbf{Robustness under Partial CT Observation.}
We test whether CSPC encourages a patient-state representation that degrades gracefully as CT evidence is removed, using uniform sparsification versus a contiguous field-of-view crop to separate the amount of retained evidence from its anatomical coverage. As shown in Figure~\ref{fig:compact_ablations}(a), we compare OmniCT-3B \cite{lin2026omnict}, HounsWorld-UND, and HounsWorld (UND+GEN), where Sparse-$r$ uniformly retains fractions $r\in\{1/2,1/4,1/8\}$ and FOV-$1/4$ retains only the contiguous central quarter, and report retention as the partial-observation report score normalized by each model's own full-observation score. CSPC is most useful as uniform sampling becomes severe: at Sparse-$1/8$ it improves retention by 3.97 points over UND and 10.66 points over OmniCT-3B, and its higher retention under Sparse-$1/4$ than FOV-$1/4$ suggests that distributed anatomical coverage is more informative than a contiguous central field of view. %Full evaluation details are given in the supplementary material.

\noindent\textbf{Transferable Reward Supervision.}\par
To test whether CSPC captures reusable clinical structure beyond HounsWorld's own architecture, we use HounsWorld as a frozen reward model to train an independently parameterized Qwen3-VL-4B~\cite{bai2025qwen3vl} with GRPO. As shown in Figure~\ref{fig:compact_ablations}(b), we report three representative settings: zero-shot inference, SFT, and GRPO augmented with HounsWorld rewards (GRPO+HW). On the random half split of the complete CT suite, GRPO+HW achieves 79.29 on closed tasks and 67.68 on open-ended QA. Its improvement over SFT, particularly on open-ended questions despite GRPO being trained with closed-form answers, suggests that CSPC supervision transfers beyond HounsWorld's own prediction head.
\noindent \textbf{Visualization and Findings}
Figure~\ref{fig:generation_visualization} illustrates the world-model view: denoising reconstructs a cleaner current observation, enhancement predicts a requested next phase, and text-and-mask conditioning simulates structured patient content. Each constructed CT stays coupled to an interpretable readout, indicating that the generative branch yields alternative observations rather than pixel-level outputs alone, and enabling controllable simulation, counterfactual interrogation, and cross-observation verification. These cases show internal semantic consistency, not independent diagnostic validity; blinded radiologist and external-reader studies remain necessary.

% \vspace{-0.8em}

\section{Conclusion}

We cast CT-centered clinical AI as world modeling: learning a shared latent representation of patient state under which readout, reconstruction, and simulation become state-dependent prediction problems rather than separate task-specific models. To study this formulation, we introduced HounsBench, a CT-centered benchmark that organizes fragmented tasks into three families, symbolic and open-ended readout, same-state reconstruction, and condition-shifted simulation, each with patient-disjoint splits and per-family protocols. On HounsBench, a single 3B HounsWorld stays competitive on readout while supporting LDCT denoising, virtual contrast enhancement, and text+mask2CT in one model, and ablations associate completion supervision with stronger, more robust readout. Our evidence is bounded to condition-specified, short-horizon observations and does not establish long-term disease evolution, treatment response, or diagnostic interchangeability of generated CT. Within this scope, HounsWorld is an evidence-bounded step from task-specific CT mappings toward unified patient-state modeling and future-observation prediction.

% Use the pre-generated bibliography so Overleaf only needs to run pdfLaTeX.
% Regenerate references.tex after editing citations or paper.bib.

\clearpage
\beginappendix
\setcounter{secnumdepth}{2}
\raggedbottom
\section{HounsBench Dataset Card}
\label{sec:dataset-card}

\hounsbench{} follows the three-family organization used in the main paper.  \readout{} evaluates symbolic and open-ended language outputs from CT.  \reconstruction{} evaluates recovery of an observation from the same patient state at \samestate{}, including dose-conditioned CT reconstruction.  \simulation{} evaluates a requested condition-shifted observation at \futurestate{}, including contrast transfer and future-state clinical text.  Text-and-mask-to-CT spans reconstruction and simulation because its target can represent either \samestate{} or \futurestate{}.  Throughout this supplement, \emph{CT completion} denotes the shared training operation rather than a fourth benchmark family.  Together these families test whether anatomy-, density-, lesion-, and enhancement-preserving objectives provide useful structural supervision for CT readout; generated volumes are not presented as independently diagnostic.

\begin{center}
  \begin{minipage}{\textwidth}
    \centering
    \includegraphics[width=0.96\textwidth]{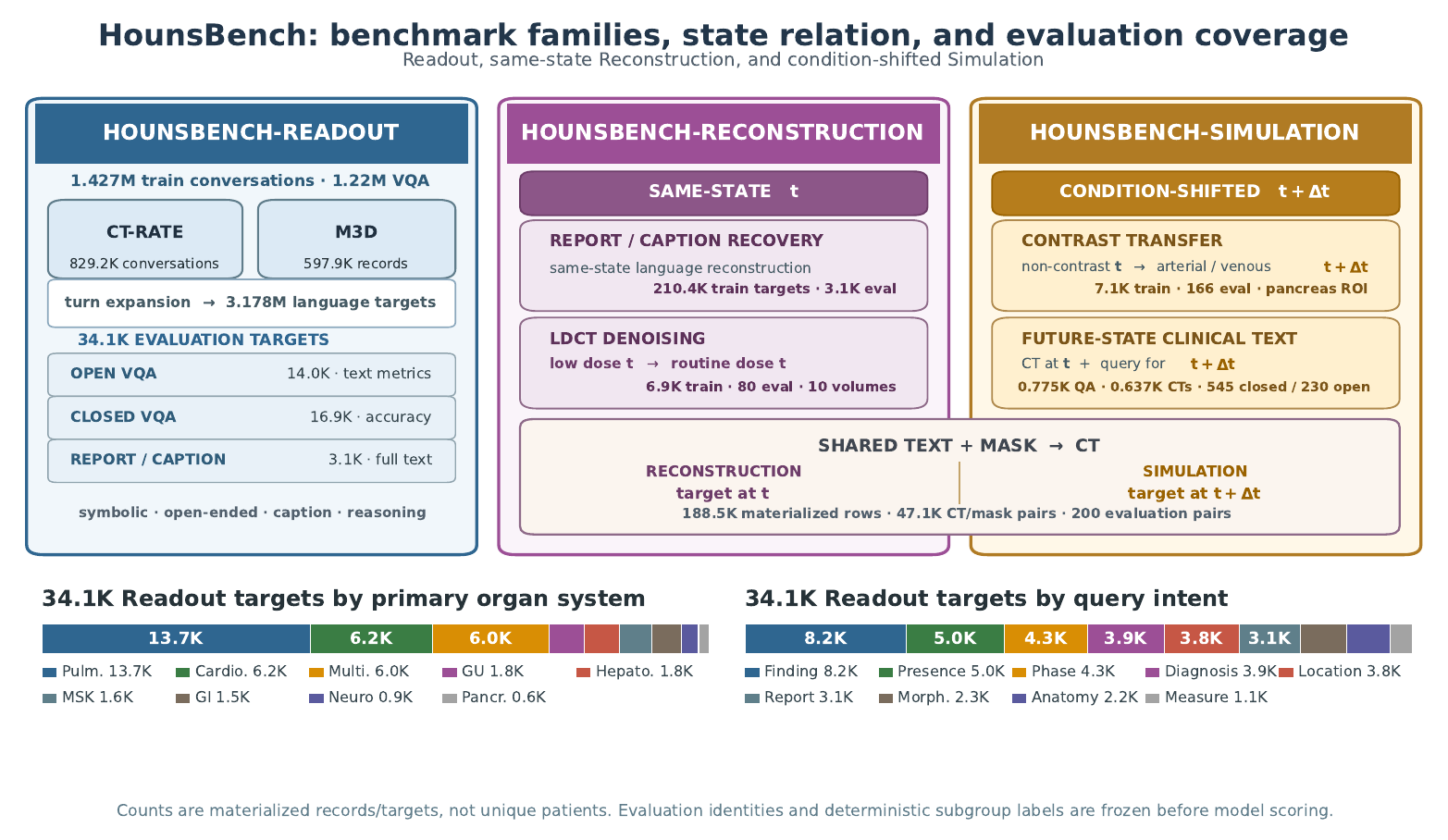}
    \captionsetup{hypcap=false}
    \captionof{figure}{Organization and scale of \hounsbench{}, aligned with the main-paper taxonomy.  The three equal-level panels summarize \readout{}, \reconstruction{} at \samestate{}, and \simulation{} at \futurestate{}.  The shared text-and-mask-to-CT rail explicitly spans reconstruction and simulation.  The lower bars report the complete 34.1K Readout evaluation distribution by primary organ system and query intent.  Values use K units for compactness; conversation, target, paired-CT, future-text-QA, and volume counts denote distinct materialization units rather than patient counts.}
    \label{fig:task-hierarchy}
  \end{minipage}
\end{center}

\subsection{Units of Counting}

We distinguish four units throughout this supplement.  A \emph{conversation record} is one stage-2 instruction item and is the unit seen by the data sampler.  A \emph{QA turn} is one human--assistant pair after multi-turn CT-RATE conversations are canonicalized.  A \emph{paired-CT item} is one input--target volume pair; several prompt variants may point to the same target volume.  A \emph{future-state clinical-text QA} is one temporally related clinical question paired with its reference answer.  This distinction reconciles the rounded main-paper counts with the audit below: the final language pool contains 1,427,089 conversation records, approximately 1.22M of which are VQA conversations, and 3,178,213 assistant targets after turn expansion, including 2,967,840 VQA turns.

\begin{table*}[!t]
\centering
\small
\setlength{\tabcolsep}{7pt}
\resizebox{\textwidth}{!}{%
\begin{tabular}{>{\bfseries}l S[table-format=4.1] S[table-format=4.1] S[table-format=3.1] S[table-format=4.1] S[table-format=3.2]}
\toprule
\rowcolor{HBBlueDark}
\tbh{Source} & {\tbh{Conversations (K)}} & {\tbh{Assistant targets (K)}} & {\tbh{Report/caption (K)}} & {\tbh{VQA turns (K)}} & {\tbh{VQA share (\%)}} \\
\midrule
\rowcolor{HBBlueLight}
CT-RATE & 829.2 & 2580.3 & 94.3 & 2486.0 & 83.76 \\
M3D & 597.9 & 597.9 & 116.1 & 481.8 & 16.24 \\
\midrule
\rowcolor{HBGrayLight}
Total & 1427.1 & 3178.2 & 210.4 & 2967.8 & 100.00 \\
\bottomrule
\end{tabular}%
}
\par\smallskip
{\footnotesize\raggedright Counts are displayed in thousands; unrounded values were retained throughout the audit.  Conversation records are sampled by the trainer, whereas assistant targets and VQA turns are counted after dialogue expansion.\par}
\caption{Stage-2 CT--language inventory, organized by counting unit.}
\label{tab:readout-train}
\end{table*}

\begin{table*}[!t]
\centering
\begin{threeparttable}
\scriptsize
\setlength{\tabcolsep}{4pt}
\begin{tabularx}{\textwidth}{>{\bfseries\raggedright\arraybackslash}p{0.14\textwidth} >{\bfseries\raggedright\arraybackslash}p{0.12\textwidth} >{\raggedright\arraybackslash}X S[table-format=3.3] >{\raggedright\arraybackslash}p{0.13\textwidth} >{\raggedright\arraybackslash}p{0.13\textwidth}}
\toprule
\rowcolor{HBPurpleDark}
\tbh{Family} & \tbh{CT task} & \tbh{State relation} & {\tbh{Train (K)}} & \tbh{Evaluation set} & \tbh{Spatial support} \\
\midrule
\rowcolor{HBPurpleLight}
Reconstruction & LDCT denoising & \makecell[l]{low-dose \samestate{} $\rightarrow$\\routine-dose \samestate{}} & 6.942 & 80 pairs / 10 volumes & Full paired volume \\
\rowcolor{HBYellowLight}
Simulation & \makecell[l]{Contrast\\transfer} & \makecell[l]{non-contrast \samestate{} $\rightarrow$\\arterial/venous \futurestate{}} & 7.101 & 166 pairs / 164 ROI-valid & \makecell[l]{Pancreas/duct/\\lesion ROI} \\
\rowcolor{HBPurpleLight}
\makecell[l]{Reconstruction/\\Simulation} & \makecell[l]{Text+mask-\\to-CT} & \makecell[l]{text + 117-label mask $\rightarrow$ CT\\at \samestate{} or \futurestate{}} & 188.452 & 200 pairs / 193 masks & \makecell[l]{Full volume /\\CT-CLIP space} \\
\bottomrule
\end{tabularx}
\begin{tablenotes}[flushleft]\footnotesize
\item Text+mask rows include multiple text views per target CT and are reported jointly at \samestate{}/\futurestate{}, matching the main-paper Reconstruction/Simulation label.  Evaluation sets are fixed, patient-disjoint from the corresponding training pools, and do not represent complete source-dataset test partitions.
\end{tablenotes}
\caption{State-conditioned CT inventory, organized by the Reconstruction/Simulation taxonomy used in the main paper.}
\label{tab:completion-inventory}
\end{threeparttable}
\end{table*}

\subsection{Scope and Source Data}

Readout supervision is derived from CT-RATE~\cite{hamamci2024ctrate} and M3D~\cite{bai2024m3d}.  CT-RATE provides chest CT volumes paired with radiology reports; M3D supplies whole-body CT image--text and VQA examples.  Our CT-language instruction construction follows the data protocol introduced with OmniCT~\cite{lin2026omnict}; Section~\ref{sec:readout-construction} states the operational transformations used in \hounsbench{} and adds a turn-level canonicalization layer required by mixed language and CT-completion training.  Report and caption targets remain in the Readout evaluation inventory because they share its language scoring protocol, while their full-state recovery is interpreted methodologically as same-state language reconstruction.

The state-conditioned CT pools use private collected cohorts and public CT data.  They comprise low-dose-to-routine-dose reconstruction at \samestate{}, non-contrast-to-arterial/venous pancreatic simulation at \futurestate{}, and text-plus-semantic-mask-conditioned CT generation spanning \samestate{}/\futurestate{}.  Anatomical masks are inferred automatically with TotalSegmentator.  All reported counts come from final materialized, patient-disjoint training and evaluation records rather than nominal source-dataset sizes.  Figure~\ref{fig:task-hierarchy} integrates source, supervision, state relation, organ-system, query-intent, and future-state clinical-text statistics into one benchmark map.

\section{HounsBench-Readout Construction}
\label{sec:readout-construction}

\subsection{Clinical Fact Decomposition}

The readout data follow the CT instruction-construction recipe of OmniCT~\cite{lin2026omnict}, while being reorganized here around patient-state readout.  For CT-RATE, a report is decomposed into report-grounded clinical facts such as the observed structure, finding or disorder, presence state, and anatomical location.  These facts are rendered into complementary question forms: open descriptions, presence questions, localization questions, disorder questions, and multiple-choice questions.  Closed questions retain an explicit answer mapping, and distractors are drawn from clinically plausible alternatives rather than arbitrary words.  Full-report targets are retained as a separate supervision type.

For M3D, the released caption and VQA records are standardized to the same volume--question--answer schema.  Short-answer and multiple-choice variants remain distinct, and caption requests are retained for CT--language alignment.  We normalize role names and visual placeholders but do not rewrite the clinical target at load time.  In both sources, a training item preserves the associated 3D CT path and its ordered dialogue.

\begin{figure*}[!t]
    \centering
    \includegraphics[width=\textwidth]{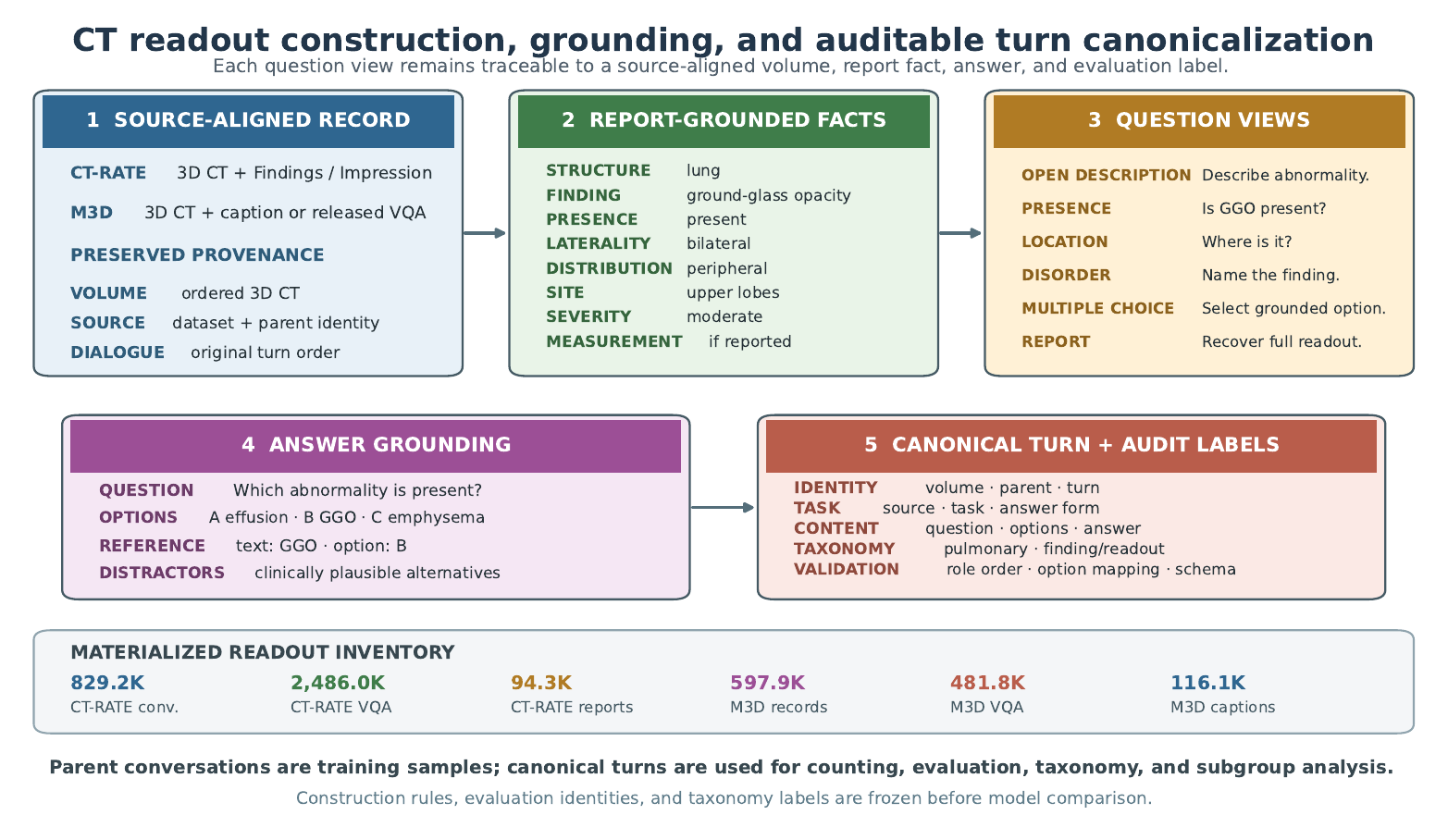}
    \caption{Readout construction and canonicalization.  A source-aligned record is decomposed into report-grounded fact fields, rendered into complementary question views, paired with grounded answers and plausible distractors, and materialized as an auditable canonical turn.  The example is schematic but uses the actual record fields.  Clinical-fact and question construction follows the OmniCT data protocol~\cite{lin2026omnict}; \hounsbench{} additionally preserves parent identities, canonical turn IDs, deterministic subgroup labels, and validation outcomes.}
    \label{fig:readout-pipeline}
\end{figure*}

Figure~\ref{fig:readout-pipeline} separates fact decomposition, answer-view construction, grounding, and canonicalization.  This keeps every answer linked to report evidence while preserving reference text, option mappings, provenance, and fixed taxonomy labels.  Path, role-order, option--answer, and schema checks run before materialization.

\subsection{Turn Canonicalization}

M3D records contain a single assistant target, whereas CT-RATE can contain several alternating question--answer pairs in one conversation.  We therefore define a canonical turn as
\begin{equation}
    t_i=(v, q_i, a_i, s, r),
\end{equation}
where $v$ is the CT volume, $q_i$ and $a_i$ are the $i$th question and answer, $s$ is the source dataset, and $r$ is the parent conversation identifier.  The parent record remains the sampling unit during training, but canonical turns are used for counting, taxonomy statistics, and subgroup evaluation.  This yields 2,485,991 CT-RATE and 481,849 M3D VQA targets (Table~\ref{tab:readout-train}).

\subsection{Evaluation Split and Task Composition}

The aligned CT-RATE and M3D evaluation set contains 34,058 language targets: 23,958 from CT-RATE and 10,100 from M3D.  Table~\ref{tab:test-inventory} gives the evaluated task blocks.  Report/caption targets and open VQA are evaluated with free-text metrics, whereas multiple-choice and presence targets use parsed accuracy.

\begin{table*}[!t]
\centering
\begin{threeparttable}
\small
\begin{tabularx}{0.96\textwidth}{>{\bfseries}l X l S[table-format=2.3] X}
\toprule
\rowcolor{HBBlueDark}
\tbh{Dataset} & \tbh{Task block} & \tbh{Answer form} & {\tbh{Samples (K)}} & \tbh{Scoring family} \\
\midrule
\multirow{6}{*}{CT-RATE}
 & Abnormality description & Open & 3.038 & Free-text metrics \\
 & Disorder description & Open & 3.038 & Free-text metrics \\
 & Location description & Open & 2.895 & Free-text metrics \\
 & Multiple choice & Closed & 8.911 & Parsed accuracy \\
 & Presence & Closed & 3.038 & Parsed accuracy \\
 & Report generation & Report & 3.038 & Free-text metrics \\
\midrule
\rowcolor{HBBlueLight}
 & Open VQA & Open & 5.000 & Free-text metrics \\
\rowcolor{HBBlueLight}
 & Multiple-choice VQA & Closed & 5.000 & Parsed accuracy \\
\rowcolor{HBBlueLight}
\multirow{-3}{*}{M3D} & Caption & Caption & 0.100 & Free-text metrics \\
\midrule
\rowcolor{HBGrayLight}
Total & Nine evaluated task blocks & Mixed & 34.058 & Reported by answer form \\
\bottomrule
\end{tabularx}
\begin{tablenotes}[flushleft]\footnotesize
\item K denotes thousands of canonical targets.  Open, closed, and report/caption targets are never collapsed into one metric because their scoring semantics differ.
\end{tablenotes}
\caption{CT-RATE and M3D evaluation inventory, grouped by answer and scoring form.}
\label{tab:test-inventory}
\end{threeparttable}
\end{table*}

\subsection{Future-State Clinical Text at \texorpdfstring{\futurestate{}}{t + Delta t}}
\label{sec:simulation-slice}

The future-state clinical-text component of \simulation{} asks a model to read the observed CT at \samestate{} and select or describe a clinically related implication at \futurestate{}: a recommended next action, an intervention or sequela timeline, a longitudinal change, an acute/chronic or recurrent state, a potential downstream consequence, or relevant history.  This component contains 775 QA rows over 637 unique CT volumes.  It complements contrast transfer and text-and-mask-conditioned CT; it is not the entirety of the Simulation family and does not claim unrestricted longitudinal trajectory generation.

\paragraph{Question-driven selection.}
We first apply deterministic lexical and template rules to the question stem, reference answer, and answer options, producing three audit priorities.  P1 requires the \emph{question itself} to invoke time/change, a clinical action, or a potential consequence at \futurestate{}.  P2 contains a generic question whose reference alone reveals a temporal state; P3 contains future wording only in a distractor.  The future-state clinical-text component keeps P1 only, so neither an answer-side cue nor a distractor-only match defines membership.  We further exclude acquisition-phase wording by itself, age-only mentions, and generic ``increased/decreased'' wording without explicit longitudinal context.  Manual audit removed four M3D modal/proper-name false positives.  P1/P2 candidates and P3 audit cases are retained separately so that the selection policy remains reversible.

\paragraph{Composition.}
Table~\ref{tab:simulation-composition} reports the task origin and intent of this \futurestate{} clinical-text component.  CT-RATE contributes 629 rows and M3D contributes 146; M3D supplies paired open and multiple-choice views of the same semantic question.  Overall, 545 rows are closed-answer and 230 are open-answer.  Intent matching is multi-label during auditing; for the mutually exclusive table, the first applicable label follows the displayed order.  Action/follow-up recommendation is the largest component (333 rows), followed by intervention/sequela timeline (198) and longitudinal change (129).

\begin{table*}[!t]
\centering
\begin{threeparttable}
\small
\setlength{\tabcolsep}{7pt}
\begin{tabularx}{0.94\textwidth}{>{\bfseries}p{0.16\textwidth} X S[table-format=3.0] S[table-format=3.2] S[table-format=3.0]}
\toprule
\rowcolor{HBBlueDark}
\makecell{\tbh{Composition}\\\tbh{view}} & \tbh{Component} & {\makecell{\tbh{QA}\\\tbh{rows}}} & {\makecell{\tbh{Share}\\\tbh{(\%)}}} & {\makecell{\tbh{Unique}\\\tbh{CTs}}} \\
\midrule
\multirow{4}{*}{Source task}
 & CT-RATE multiple choice & 464 & 59.87 & 447 \\
 & CT-RATE localization & 157 & 20.26 & 157 \\
 & CT-RATE presence & 8 & 1.03 & 8 \\
 & M3D paired open/closed VQA & 146 & 18.84 & 63 \\
\midrule
\rowcolor{HBBlueLight}
 & Closed answer & 545 & 70.32 & 515 \\
\rowcolor{HBBlueLight}
\multirow{-2}{*}{\makecell[c]{Answer\\form}} & Open answer & 230 & 29.68 & 220 \\
\midrule
\multirow{6}{*}{\makecell[l]{Primary\\simulation\\intent}}
 & Action / follow-up recommendation & 333 & 42.97 & 332 \\
 & Intervention or sequela timeline & 198 & 25.55 & 172 \\
 & Longitudinal change & 129 & 16.65 & 95 \\
 & Clinical temporal state & 71 & 9.16 & 34 \\
 & Potential downstream consequence & 39 & 5.03 & 20 \\
 & History context & 5 & 0.65 & 5 \\
\midrule
\rowcolor{HBGrayLight}
Total & Future-state clinical text & 775 & 100.00 & 637 \\
\bottomrule
\end{tabularx}
\begin{tablenotes}[flushleft]\footnotesize
\item Unique-CT counts are unions within each row and are not additive across rows because one volume can support multiple QA views.  All matched intent labels are retained during auditing; the primary label is used only to make the six intent rows mutually exclusive.
\end{tablenotes}
\caption{Selection and composition of the \futurestate{} clinical-text component within \simulation{}.}
\label{tab:simulation-composition}
\end{threeparttable}
\end{table*}

\subsection{Organ-System and Query-Intent Taxonomies}

To expose coverage beyond aggregate task labels, every VQA turn receives one primary organ-system label and one query-intent label.  Organ labels are assigned from question evidence with reference-answer fallback.  The nine organ systems are neuro/head--neck, pulmonary/pleural, cardiovascular/mediastinal, hepatobiliary, pancreas/spleen, genitourinary/adrenal, gastrointestinal, musculoskeletal/soft tissue, and multisystem/other.  Query intent uses nine mutually exclusive labels: acquisition/phase, measurement/counting, presence/state, localization/laterality, diagnosis/etiology, morphology/attenuation, anatomy/organ identification, finding/readout, and report/caption.

The mapping is deterministic rather than learned.  Source-specific CT-RATE task labels take precedence for abnormality, disorder, location, presence, and report questions; the remaining cases are resolved by an ordered lexical rule set.  We retain a primary label for compact plots, even when a case mentions more than one organ.  Patterns, precedence, subgroup-size thresholds, and canonical labels are frozen before scoring, preventing post-hoc relabeling based on model performance.

The evaluation set is chest-heavy: 40.17\% of targets receive the pulmonary/pleural label and 18.27\% the cardiovascular/mediastinal label.  The most common query intents are finding/readout (24.11\%), presence/state (14.73\%), acquisition/phase (12.52\%), diagnosis/etiology (11.45\%), and localization/laterality (11.23\%).  Figure~\ref{fig:task-hierarchy} visualizes the complete distribution in K units, including long-tail abdominal and musculoskeletal systems.

\section{HounsBench Reconstruction and Simulation Construction}
\label{sec:completion-construction}

\subsection{Low-Dose Denoising}

Low-dose denoising belongs to \reconstruction{} because it recovers a spatially matched routine/reconstructed-dose observation of the same patient state at \samestate{}.  The training pool contains 6,942 pairs, and the evaluation set contains 80 pairs from ten cases.  This task emphasizes density recovery and preservation of anatomical boundaries under acquisition noise; it is used as auxiliary supervision rather than as evidence that generated images should replace diagnostic reconstruction.

\subsection{Contrast Transfer}

Contrast transfer belongs to \simulation{}: it maps a non-contrast pancreatic CT patch observed at \samestate{} to the registered arterial or venous observation requested at \futurestate{} for the same patient and spatial region.  The 7,101-pair training pool contains 3,550 arterial and 3,551 venous targets.  The evaluation set contains 83 non-contrast patches, each paired with both phases, for 166 pairs.  Holding patient and patch position fixed makes this a bounded condition-shifted prediction of phase-dependent parenchymal and lesion appearance rather than unrestricted disease forecasting.

\begin{figure*}[!t]
    \centering
    \includegraphics[width=0.86\textwidth]{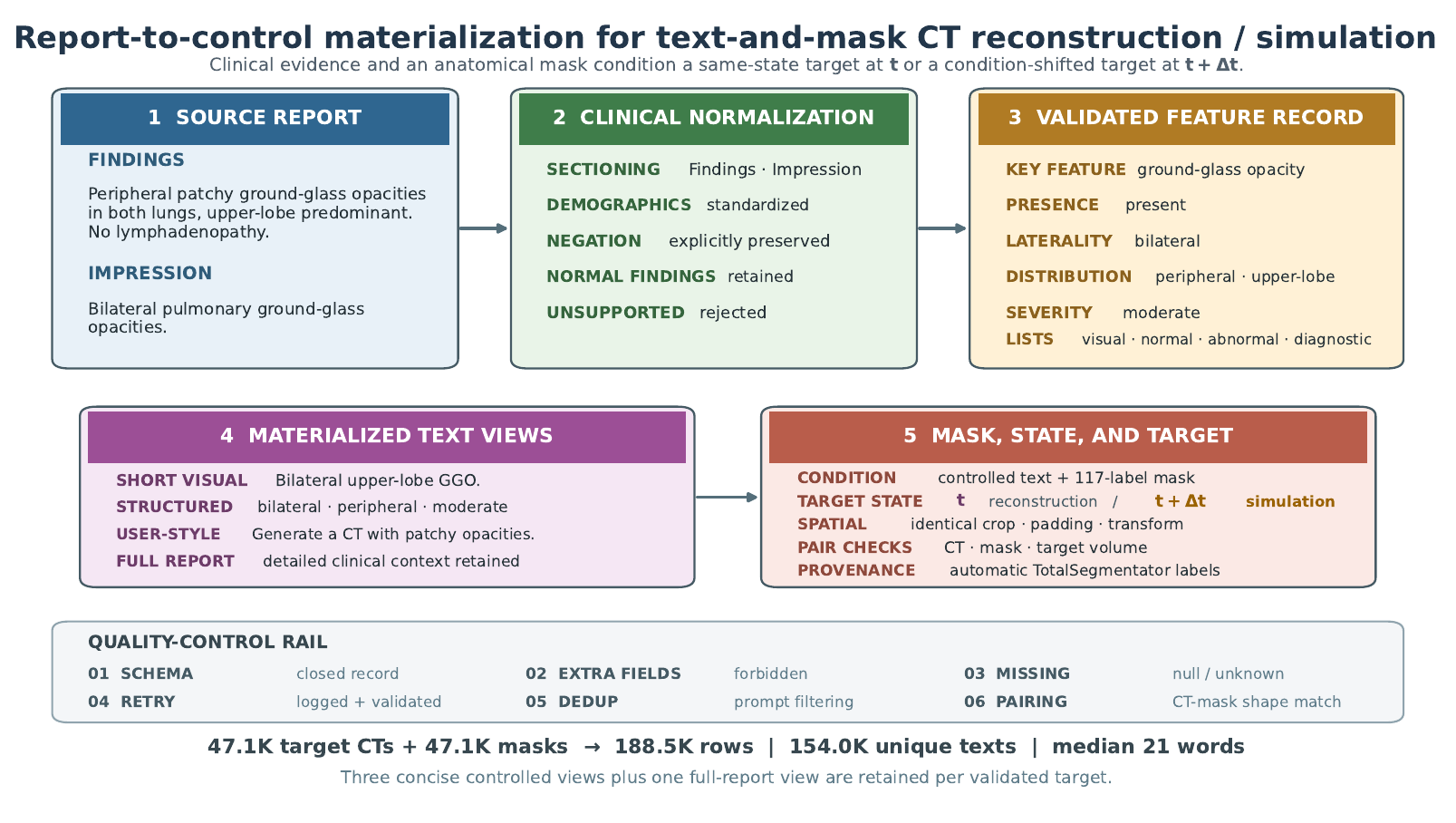}
    \caption{Report-to-control materialization for text-and-mask-conditioned CT reconstruction/simulation.  The upper path shows report sectioning, normalization, and a concrete structured-feature example; the lower path shows four retained text views, spatial pairing with a 117-label TotalSegmentator mask, the highlighted target-state choice \samestate{}/\futurestate{}, and the validation rail.  Counts at the bottom distinguish target CTs, masks, materialized rows, and unique texts.}
    \label{fig:prompt-pipeline}
\end{figure*}

\subsection{Text-and-Mask-to-CT}

Text-and-mask-to-CT spans \reconstruction{} and \simulation{} because the same interface can instantiate a target observation at \samestate{} or a requested condition at \futurestate{}.  Consistent with the main paper, its inventory and metrics are reported jointly as Reconstruction/Simulation.  The rounded 188K count in the main paper denotes materialized prompt-conditioned training examples; the underlying inventory contains 47,113 unique target CT/mask pairs.  The text condition is not the untouched CT-RATE report: we use the pipeline in Figure~\ref{fig:prompt-pipeline} to convert variable free text into controllable patient-state descriptions.

\paragraph{Report sectioning and normalization.}
We first separate Findings and Impression and normalize demographic expressions without changing the clinical target.  An instruction model then converts the report into a constrained clinical representation that separates salient visual features, explicitly normal observations, abnormal observations, and optional diagnostic labels.  The controlled vocabulary covers ground-glass opacity, consolidation, nodules, emphysema, fibrosis/scarring, bronchiectasis, pleural effusion, lymphadenopathy, heart size, and overall lung aeration.  Feature attributes encode presence, laterality, spatial distribution, severity, and the largest nodule diameter when explicitly stated.  Diagnostic labels cover normal examination, pneumonia, COVID-19 pneumonia, and COPD-related change without inferring an unreported diagnosis.

The representation must pass a closed clinical schema: unsupported fields are rejected, unmentioned observations remain absent or unknown, and failed outputs are regenerated and checked again.  This prevents unsupported free-form prose from entering the conditioning text.  The representative example in Figure~\ref{fig:prompt-pipeline} illustrates how bilateral, peripheral, upper-lobe-predominant ground-glass opacity becomes a controllable combination of presence, laterality, distribution, and severity.

\paragraph{Control-focused prompt materialization.}
We rank clinically informative abnormal features, select at most two primary controls, and retain a limited number of secondary or normal constraints.  A deterministic renderer produces short visual, medium visual, user-style, structured, and diagnosis-enhanced candidates.  It orders distribution tags, combines normalized spatial modifiers with natural key-visual phrases, removes redundant minor findings, and adds a normal constraint only when explicitly supported.  Three concise prompts are selected for each target, and one full-report-conditioned prompt is retained to preserve detailed language.  Thus each validated target CT/mask normally contributes four training conditions.

The final materialization contains 188,452 rows associated with 47,113 target CT volumes and 47,113 aligned masks, and includes 154,026 unique prompt strings.  Prompts have a median length of 21 words; the mean is 75.00 words because the fourth view retains the complete report.  These paired concise/full conditions deliberately expose the model to both direct control language and clinically rich context.

\paragraph{Mask provenance.}
All semantic conditions are automatically inferred by TotalSegmentator~\cite{wasserthal2023totalsegmentator} and transformed with the paired CT.  Our 117-label mapping covers the foreground anatomical structures used for conditioning.  These masks are model-generated anatomical conditions, not radiologist-drawn ground truth.  Their automatic provenance should be considered when interpreting mask-conditioned generation results.

\begin{figure*}[!t]
    \centering
    \includegraphics[width=0.70\textwidth]{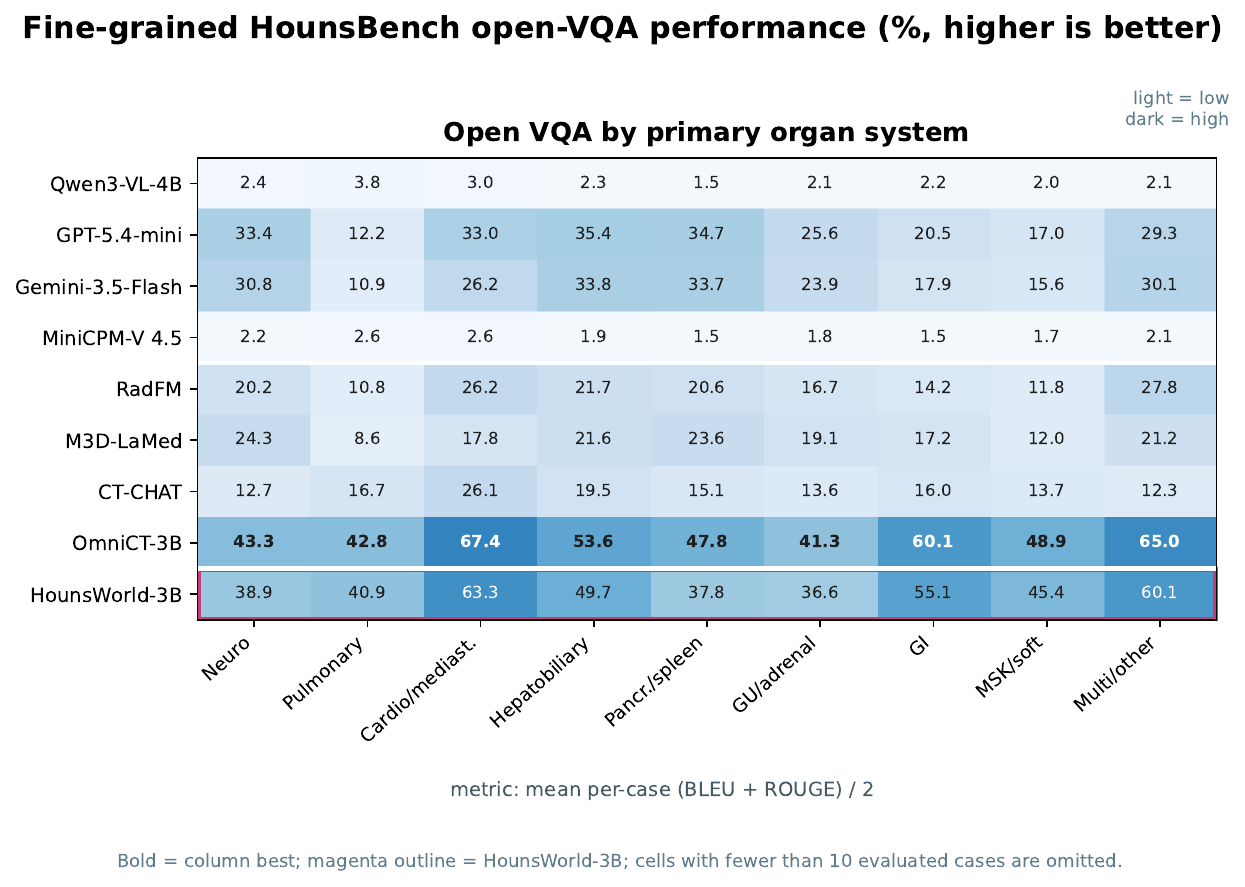}
    \caption{Fine-grained open-VQA performance grouped by primary organ system and scored with per-case $(\text{BLEU}+\text{ROUGE})/2$.  Bold cells are column maxima; the outlined row is \hounsworld{}.  Values are percentages, and groups with fewer than ten evaluated examples are omitted.  These descriptive diagnostics are not the main-paper composite.}
    \label{fig:subgroup-results}
\end{figure*}

\section{Evaluation Protocol and Fine-Grained Readout}
\label{sec:evaluation}

\subsection{Readout Metrics}

Closed VQA uses CT-FAIR parsed accuracy.  The parser accepts explicit answer letters and conservative option-text matches, handles yes/no semantics for presence questions, and counts unparseable predictions as incorrect when the reference is valid.  Open VQA and report/caption targets use BLEU~\cite{papineni2002bleu}, ROUGE~\cite{lin2004rouge}, RadGraph-XL~\cite{delbrouck2024radgraphxl}, and BioBERTScore~\cite{zhang2020bertscore,lee2020biobert}.  The main-paper open score is
\begin{equation}
S_{\mathrm{open}}=0.1S_{\mathrm{BLEU}}+0.1S_{\mathrm{ROUGE}}+0.4S_{\mathrm{RG}}+0.4S_{\mathrm{BioBERT}}.
\end{equation}

The fine-grained figure below serves a different diagnostic purpose.  Because per-case RadGraph-XL and BioBERTScore values were not available uniformly for every external API model, its open-VQA cells use the lexical score $(S_{\mathrm{BLEU}}+S_{\mathrm{ROUGE}})/2$.  This organ-level view is reported as a descriptive analysis rather than as the main-paper composite.

The \futurestate{} clinical-text component of \simulation{} in Section~\ref{sec:simulation-slice} inherits this answer-form separation: its 545 closed questions use parsed accuracy, while its 230 open questions use $(S_{\mathrm{BLEU}}+S_{\mathrm{ROUGE}})/2$.  We report the two values independently because a row-weighted average would conflate exact option selection with lexical overlap.

Figure~\ref{fig:subgroup-results} shows that the organ taxonomy exposes substantial variation hidden by task-level averages.  For \hounsworld{}, the strongest open-overlap categories are cardiovascular/mediastinal (63.3), multisystem/other (60.1), gastrointestinal (55.1), and hepatobiliary (49.7).  We treat these values as descriptive subgroup diagnostics, not as independent clinical validation.

\subsection{Reconstruction and Simulation Metrics}

All paired CT arrays are clipped to $[-1000,1000]$ HU and linearly mapped to $[0,1]$.  Reconstruction by LDCT denoising at \samestate{} reports volume PSNR and normalized RMSE together with slice-averaged SSIM~\cite{wang2004ssim}.  Simulation by contrast transfer at \futurestate{} reports the same metrics inside a pancreas-centered semantic ROI.  The ROI is the union of pancreas, pancreatic duct when retained for PDAC cases, and pancreatic lesion labels.  PSNR/RMSE operate on all ROI voxels; ROI-SSIM averages pixels whose full $11\times11$ support lies inside an eroded semantic mask.  Of the 166 contrast-transfer evaluation pairs, 164 contain a valid ROI.

Text-and-mask-to-CT reports PSNR, LPIPS~\cite{zhang2018lpips}, and CT-KID.  LPIPS uses an AlexNet backbone on axial slices replicated to three channels, followed by volume- and sample-level averaging.  CT-KID is a KID-style unbiased cubic-kernel MMD in a 512-dimensional CT-CLIP embedding space.  It uses 100 random subsets of 100 generated and reference volumes.  For readability, we report $10^3\!\times$ CT-KID, matching the numerical scale used in the main paper; the underlying estimator is not bounded to $[0,1]$.

\begin{table*}[!t]
\centering
\begin{threeparttable}
\footnotesize
\setlength{\tabcolsep}{4pt}
\begin{tabularx}{0.97\textwidth}{>{\centering\arraybackslash\bfseries}p{0.15\textwidth} >{\raggedright\arraybackslash}p{0.14\textwidth} X >{\centering\arraybackslash}p{0.09\textwidth} >{\raggedright\arraybackslash}p{0.17\textwidth}}
\toprule
\rowcolor{HBBlueDark}
\tbh{Branch} & \tbh{Task} & \tbh{Primary measures} & \tbh{Direction} & \tbh{Scoring support} \\
\midrule
\rowcolor{HBBlueLight}
 & Closed VQA & Parsed accuracy & $\uparrow$ & Valid closed-form targets \\
\rowcolor{HBBlueLight}
\multirow{-2}{*}{Readout} & Open/report & \makecell[l]{BLEU; ROUGE; RadGraph-XL;\\BioBERTScore} & $\uparrow$ & Per-target text \\
\midrule
\rowcolor{HBPurpleLight}
Reconstruction & \makecell[l]{LDCT denoising\\\samestate{}} & PSNR; SSIM; normalized RMSE & $\uparrow,\uparrow,\downarrow$ & Full paired volume \\
\midrule
\rowcolor{HBYellowLight}
 & \makecell[l]{Contrast transfer\\\futurestate{}} & ROI-PSNR; ROI-SSIM; ROI-RMSE & $\uparrow,\uparrow,\downarrow$ & \makecell[l]{Pancreas/duct/\\lesion union} \\
\rowcolor{HBYellowLight}
 & \makecell[l]{Closed VQA\\\futurestate{}} & Parsed accuracy & $\uparrow$ & 545 closed-form targets \\
\rowcolor{HBYellowLight}
\multirow{-3}{*}{Simulation} & \makecell[l]{Open VQA\\\futurestate{}} & BLEU; ROUGE & $\uparrow$ & 230 free-text targets \\
\midrule
\rowcolor{HBPurpleLight}
\makecell[c]{Reconstruction/\\Simulation} & \makecell[l]{Text+mask-\\to-CT\\\samestate{}/\futurestate{}} & PSNR; $10^3\!\times$ CT-KID; LPIPS & $\uparrow,\downarrow,\downarrow$ & \makecell[l]{Full volume /\\CT-CLIP space} \\
\bottomrule
\end{tabularx}
\begin{tablenotes}[flushleft]\footnotesize
\item Arrows follow the measure order within each row.  Readout and future-state clinical-text answer forms remain separate.  CT completion is the shared prediction operator; the displayed family labels follow the same-state \samestate{} versus condition-shifted \futurestate{} organization used in the main paper.
\end{tablenotes}
\caption{Evaluation measures, direction, spatial support, and main-paper HounsBench family assignment.}
\label{tab:metrics}
\end{threeparttable}
\end{table*}

\section{Qualitative Visualization}
\label{sec:qualitative-visualization}

\makeatletter
\setlength{\@dblfptop}{0pt}
\makeatother

We visualize two cases for each of two state-conditioned CT tasks.  Non-contrast-to-contrast transfer is a \simulation{} task that predicts a requested observation at \futurestate{}; text-and-mask-to-CT spans \reconstruction{}/\simulation{} at \samestate{}/\futurestate{}.  Figures~\ref{fig:contrast-case1}--\ref{fig:tm2ct-case2} show the conditioning input, baseline outputs, the \hounsworld{} output, and paired ground truth over the same axial positions.  These examples expose whether a method changes only global appearance or preserves condition-specific anatomy and tissue behavior across the volume.

\subsection{Contrast Transfer}

\begin{figure*}[!t]
    \centering
    \includegraphics[width=0.81\textwidth]{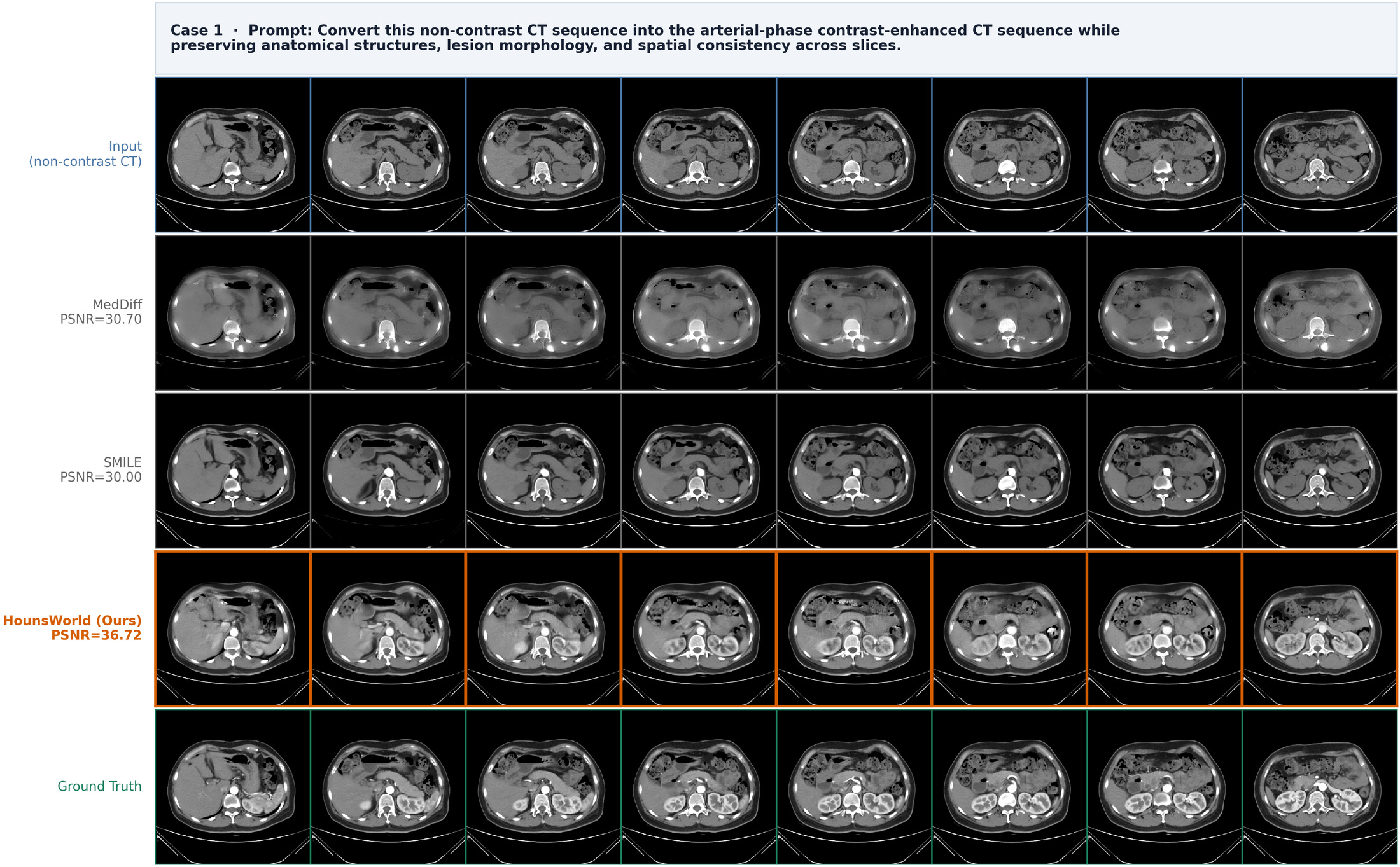}
    \caption{Arterial-phase \simulation{} at \futurestate{}, Case~1.  Rows show the non-contrast input at \samestate{}, MedDiff, SMILE, \hounsworld{}, and the paired arterial-phase ground truth.  MedDiff produces diffuse intensity amplification, while SMILE enhances the arterial lumen but under-enhances liver and kidney parenchyma.  Displayed PSNR values are volume-level ROI-PSNR.}
    \label{fig:contrast-case1}
    \includegraphics[width=0.81\textwidth]{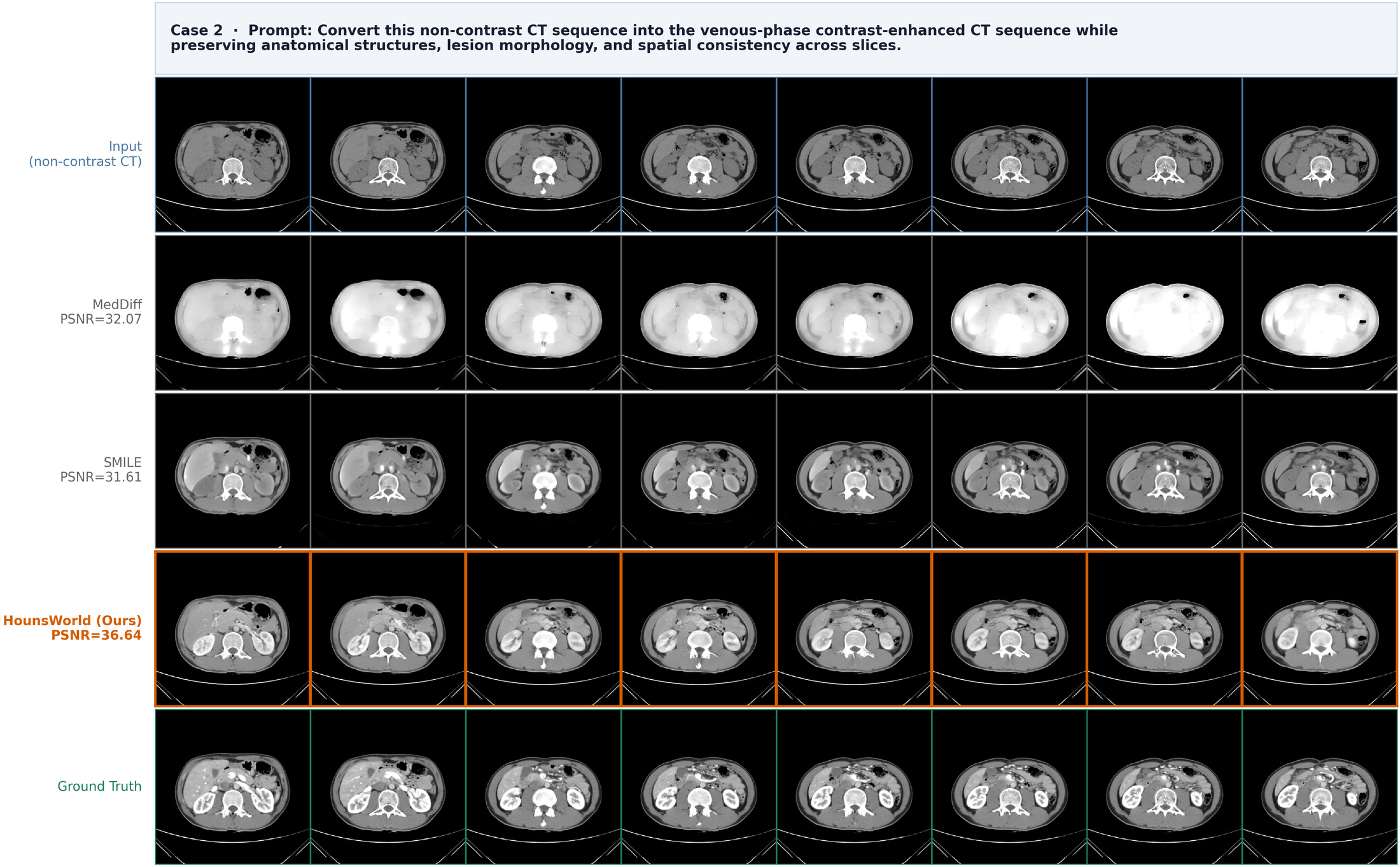}
    \caption{Venous-phase \simulation{} at \futurestate{}, Case~2.  Rows follow Figure~\ref{fig:contrast-case1}.  MedDiff yields a spatially diffuse, nearly saturated response; SMILE partially enhances the liver but largely misses the renal response.  \hounsworld{} produces more coordinated enhancement across vascular and parenchymal structures.  Displayed PSNR values are volume-level ROI-PSNR.}
    \label{fig:contrast-case2}
\end{figure*}

Figures~\ref{fig:contrast-case1} and~\ref{fig:contrast-case2} visualize bounded condition-shifted prediction from non-contrast \samestate{} to arterial/venous \futurestate{}.  They reveal that visually brighter output does not necessarily represent physically meaningful contrast transfer.  MedDiff applies a broad, smooth intensity change whose enhancement is poorly localized to vessels or perfused parenchyma.  This is especially apparent in Case~2, where large abdominal regions become nearly saturated without the organ-dependent enhancement pattern expected from a venous acquisition.  SMILE better preserves the input geometry, but its enhancement remains incomplete.  In Case~1 it enhances the arterial lumen while leaving hepatic and renal parenchyma close to the non-contrast input.  In Case~2 the liver receives some enhancement, whereas renal enhancement is largely neglected.

By comparison, \hounsworld{} couples vascular enhancement with changes in liver and kidney appearance and maintains these changes across adjacent slices.  The renal enhancement pattern and the transition of abdominal structures are consequently closer to the paired target.  The examples also expose a remaining limitation: fine parenchymal texture and small vascular detail are smoother than in the ground truth.  Thus, the principal qualitative advantage is coordinated, spatially persistent phase transfer rather than simple intensity amplification.

\subsection{Text-and-Mask-to-CT Generation}

\begin{figure*}[!t]
    \centering
    \includegraphics[width=0.81\textwidth]{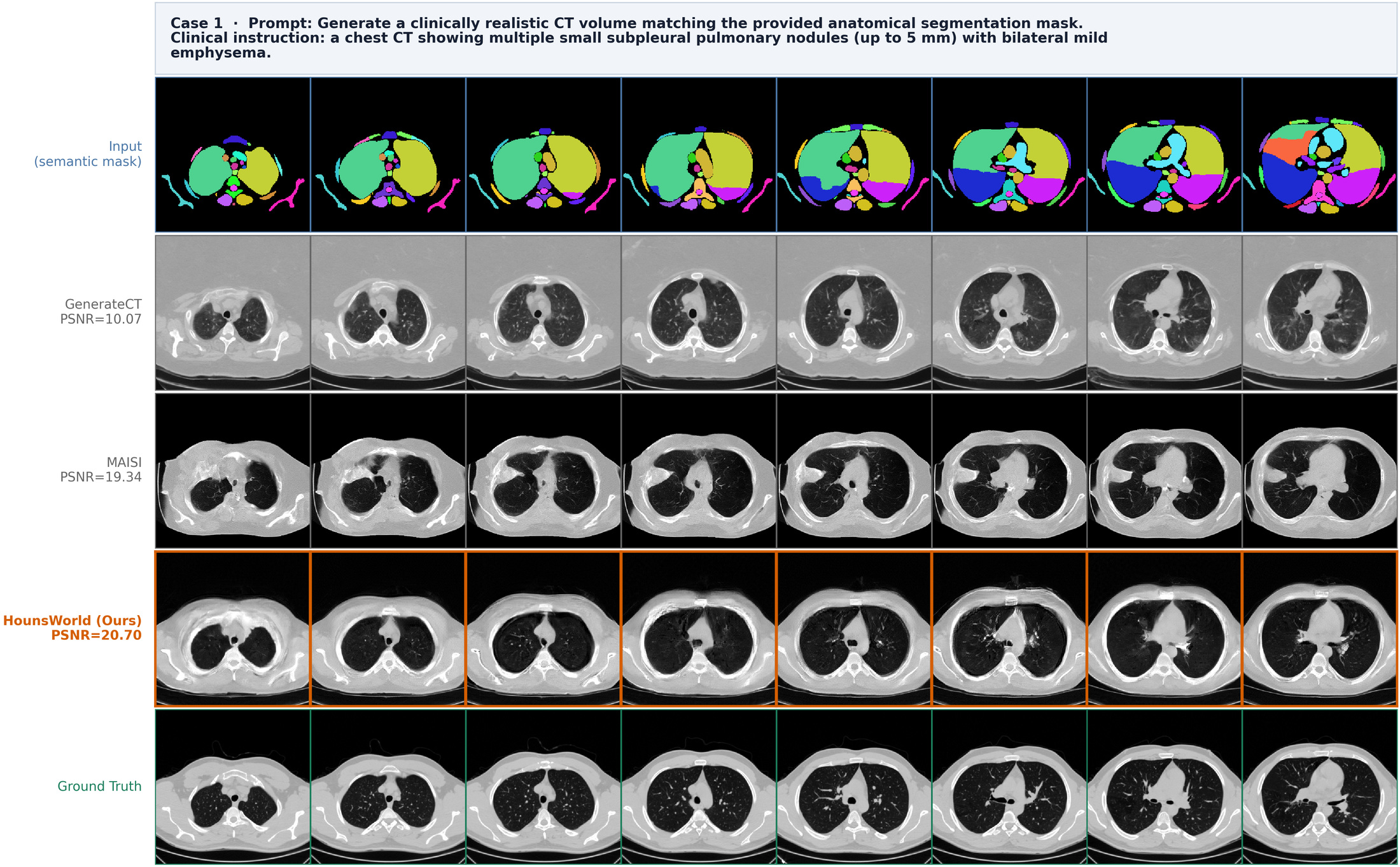}
    \caption{Text-and-mask-to-CT Reconstruction/Simulation at \samestate{}/\futurestate{}, Case~1.  GenerateCT receives \emph{text only}, MAISI receives the \emph{semantic mask only}, and \hounsworld{} receives \emph{both text and mask}.  GenerateCT produces plausible lung texture but weak case-specific spatial control; MAISI introduces irregular pulmonary noise, whereas the joint condition better preserves anatomical organization.  The bottom row is the paired ground truth.}
    \label{fig:tm2ct-case1}
    \includegraphics[width=0.81\textwidth]{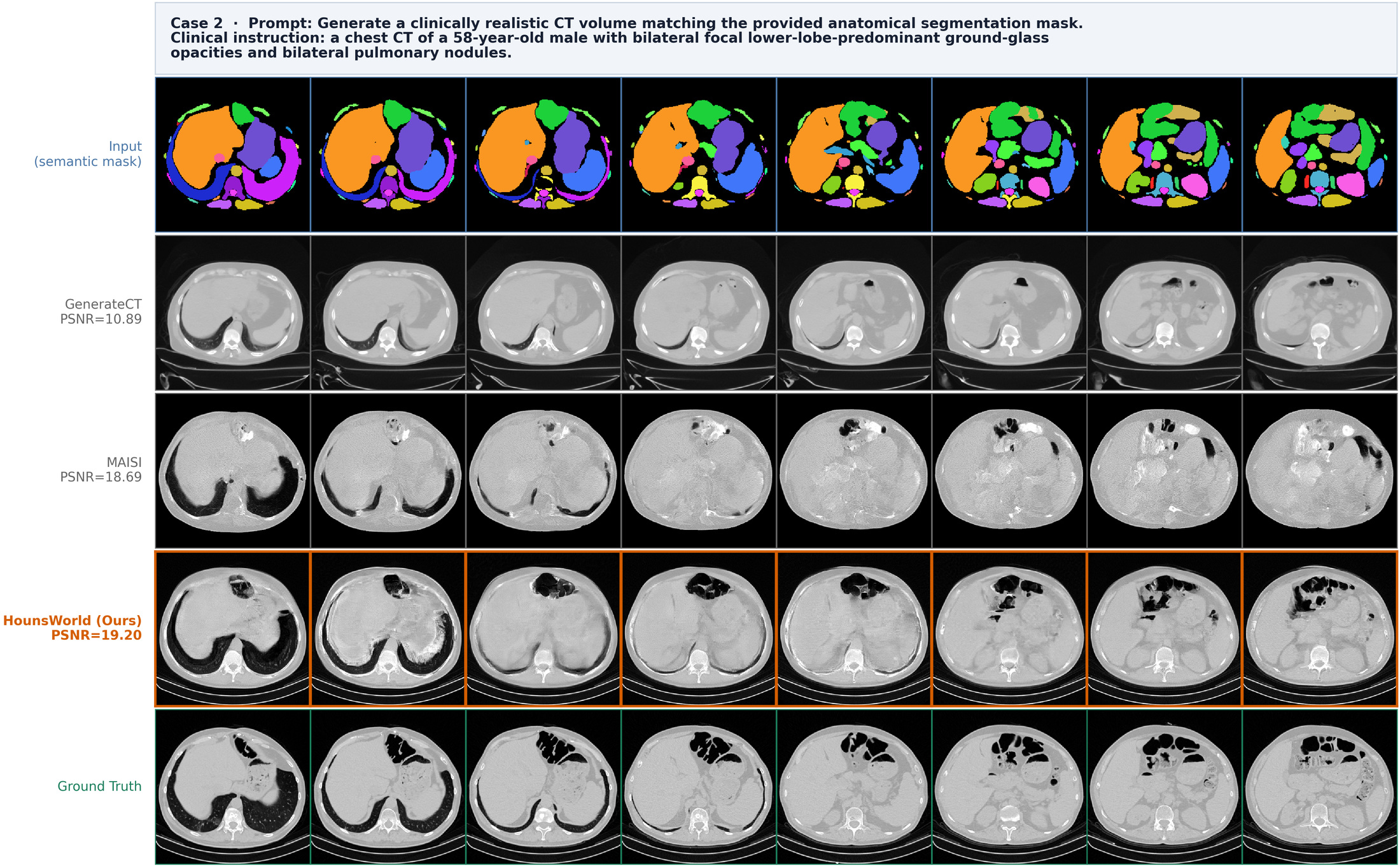}
    \caption{Text-and-mask-to-CT Reconstruction/Simulation at \samestate{}/\futurestate{}, Case~2.  GenerateCT receives \emph{text only}, MAISI receives the \emph{semantic mask only}, and \hounsworld{} receives \emph{both text and mask}.  The text-only output does not follow the case-specific organ layout, while the mask-only output retains weak organ boundaries and texture.  Joint conditioning provides more consistent anatomical progression, although fine detail remains below the paired ground truth.}
    \label{fig:tm2ct-case2}
\end{figure*}

Figures~\ref{fig:tm2ct-case1} and~\ref{fig:tm2ct-case2} show substantial case-to-case variation for GenerateCT.  In Case~1 it synthesizes relatively plausible pulmonary texture, but the overall CT anatomy and spatial semantics are weakly controlled with respect to the supplied mask and target.  The abdominal sequence in Case~2 is smoother and more coherent internally, yet its body geometry and organ arrangement still do not follow the case-specific spatial condition.  This behavior is consistent with a text-only interface: textual findings can influence appearance, but they do not specify the voxel-level anatomical layout.

MAISI receives the semantic mask and therefore has access to coarse spatial structure, but its fine appearance control remains limited.  Case~1 contains irregular, non-physical pulmonary noise and peripheral texture that are not supported by the requested state.  In Case~2, organ topology, tissue boundaries, and internal texture are poorly resolved despite the mask condition.  Moreover, a mask-only interface cannot express the textual abnormality controls available to the other methods.  \hounsworld{} combines the complementary conditions: its organ layout more consistently follows the semantic mask while the text controls the requested patient state.  Across the displayed slices, it also exhibits more coherent anatomical progression.  Nevertheless, its fine lung texture and small intra-organ structures remain smoother than the paired ground truth, indicating that spatial-semantic control improves before full high-frequency fidelity is reached.

\makeatletter
\setlength{\@dblfptop}{0pt plus 1fil}
\makeatother

\FloatBarrier
\section{Minimal Architecture and Training Details}
\label{sec:implementation}

\subsection{Variable-Depth CT Handling}

The semantic and generative routes impose different depth congruence constraints.  For an input of depth $D$, the ViT route requires a multiple of three slices,
\begin{equation}
    D_{\mathrm{ViT}}=3\left\lceil D/3 \right\rceil,
\end{equation}
whereas the VAE route requires a depth of the form $1+4X$,
\begin{equation}
    D_{\mathrm{VAE}}=1+4\left\lceil (D-1)/4 \right\rceil.
\end{equation}
If the input does not meet the corresponding constraint, boundary slices are padded along the cranio-caudal axis to the next admissible depth.  Source CT, target CT, and semantic mask receive the same padding.  Padding is removed after decoding before computing paired metrics.

\subsection{Two-Stage Optimization}

Training samples CT--language supervision, LDCT denoising, contrast transfer, and text-and-mask-to-CT with configured 60/10/10/20 weights, not guaranteed observed proportions.  Both stages use language CE weight 1.0 and CT flow-matching/MSE weight 0.25; Table~\ref{tab:training} summarizes the remaining configuration.

\begin{table}[!htbp]
\centering
\begin{threeparttable}
\scriptsize
\setlength{\tabcolsep}{2.5pt}
\begin{tabularx}{\columnwidth}{>{\bfseries\raggedright\arraybackslash}p{0.19\columnwidth} >{\raggedright\arraybackslash}X >{\raggedright\arraybackslash}X}
\toprule
\rowcolor{HBBlueDark}
 & \tbh{Stage 1: CT alignment} & \tbh{Stage 2: joint adaptation} \\
\midrule
\rowcolor{HBBlueLight}
Language pool & 163.2K report/caption conversations & 1,427.1K CT--language conversations \\
Sampling mixture & CT--language alignment & 60/10/10/20 task weights \\
\midrule
\rowcolor{HBGrayLight}
Trainable modules & Three bottleneck-128 residual CT adapters & ViT/merger, token embeddings, both Qwen2-MoT routes, LM head, and CT adapters \\
Frozen modules & Visual encoder, language model, Wan VAE, and Lance latent connectors & Wan VAE and pretrained Lance latent connectors \\
\midrule
\rowcolor{HBBlueLight}
Learning rate & $2\times10^{-4}$; 100 warmup steps & $2\times10^{-5}$; 200 warmup steps \\
Loss weights & CE $=1.0$; CT flow-matching/MSE $=0.25$ & CE $=1.0$; CT flow-matching/MSE $=0.25$ \\
Precision / parallelism & bfloat16 FSDP on 16 GPUs & bfloat16 FSDP on 16 GPUs \\
Final step & 1,170 & 9,943 \\
\bottomrule
\end{tabularx}
\begin{tablenotes}[flushleft]\scriptsize
\item Sampling ratios are configured sampler weights, not guarantees that every unique item is observed in that exact proportion.  Both stages stop after their first language-pool cycle.
\end{tablenotes}
\caption{Released two-stage training configuration, grouped by data, modules, and optimization.}
\label{tab:training}
\end{threeparttable}
\end{table}

Training uses bfloat16 FSDP on 16 GPUs with a constant learning-rate schedule.  Stage 2 initializes from the independent Stage-1 checkpoint; ``joint adaptation'' does not update the frozen Wan VAE.

\FloatBarrier
\section{Quality Control, Reproducibility, and Limitations}
\label{sec:governance}
\enlargethispage{2\baselineskip}

\paragraph{Schema and pairing checks.}
Every readout item is required to contain an available CT volume and an alternating human--assistant sequence.  Reconstruction/Simulation CT pairs are checked for source/target shape agreement, and the same spatial transformation is replayed for CT and masks.  Structured report conversion uses schema validation and retry logging.  Aggregate counts are regenerated from the frozen benchmark state under the same deterministic rules.

\paragraph{Automatic labels.}
TotalSegmentator masks are automatic pseudo-labels.  They enable scalable spatial conditioning but can inherit segmentation errors, especially for small or abnormal structures.  The organ/intention taxonomy is also automatic and rule-based.  It is useful for coverage audits and deterministic subgroup plots, but it is not a substitute for radiologist adjudication.

\paragraph{Metric interpretation.}
Text-overlap measures may favor concise answers and do not fully measure clinical correctness.  RadGraph-XL and BioBERTScore improve semantic sensitivity but can behave poorly on extremely short answers.  Completion fidelity metrics measure agreement with paired targets; they do not establish diagnostic interchangeability.  CT-KID is a finite-sample MMD estimator in CT-CLIP space and should be interpreted comparatively under an identical protocol.

\paragraph{Privacy and release.}
The supplementary statistics expose only aggregate counts, deterministic category labels, and benchmark identifiers.  They do not release protected patient metadata.  Dataset access and redistribution remain governed by the licenses and use agreements of CT-RATE, M3D, and the contributing clinical sources.

\paragraph{Reproducibility protocol.}
Canonical evaluation identities, future-state clinical-text selection rules, taxonomy precedence, subgroup-size thresholds, and scoring families are fixed before model comparison.  Open and closed subgroup results are maintained separately throughout aggregation to prevent accidental cross-scale averaging.

\end{document}